\documentclass[journal]{IEEEtran}

\usepackage{epsfig}
\usepackage{amssymb}
\usepackage{amsthm}
\usepackage{graphicx}
\usepackage{xcolor}
\usepackage{boldline,multirow}
\usepackage{hyperref}
\usepackage{amsmath}

\begin{document}

\title{Efficient Perception, Planning, and Control Algorithm for Vision-Based Automated Vehicles}

\author{Der-Hau Lee
\thanks{The author was with the Department of Electrophysics,  National Yang Ming Chiao Tung University, Hsinchu 300, Taiwan. e-mail: derhaulee@gmail.com.}% <-this % stops a space
}

% make the title area
\maketitle

\begin{abstract}
Autonomous vehicles have limited computational resources and thus require efficient control systems. The cost and size of sensors have limited the development of self-driving cars. To overcome these restrictions, this study proposes an efficient framework for the operation of vision-based automatic vehicles; the framework requires only a monocular camera and a few inexpensive radars. The proposed algorithm comprises a multi-task UNet (MTUNet) network for extracting image features and constrained iterative linear quadratic regulator (CILQR) and vision predictive control (VPC) modules for rapid motion planning and control. MTUNet is designed to simultaneously solve lane line segmentation, the ego vehicle’s heading angle regression, road type classification, and traffic object detection tasks at approximately 40 FPS  for 228 $\times$ 228 pixel RGB input images. The CILQR controllers then use the MTUNet outputs and radar data as inputs to produce driving commands for lateral and longitudinal vehicle guidance within only 1 ms. In particular, the VPC algorithm is included to reduce steering command latency to below actuator latency, preventing performance degradation during tight turns. The VPC algorithm uses road curvature data from MTUNet to estimate the appropriate correction for the current steering angle at a look-ahead point to adjust the turning amount. The inclusion of the VPC algorithm in a VPC-CILQR controller leads to higher performance on curvy roads than the use of CILQR alone. Our experiments demonstrate that the proposed autonomous driving system, which does not require high-definition maps, can be applied in current autonomous vehicles.
\end{abstract}

% Note that keywords are not normally used for peerreview papers.

\section{Introduction}

\IEEEPARstart{T}{he} use of deep neural network (DNN) techniques in intelligent vehicles has expedited the development of self-driving vehicles in research and industry. Self-driving cars can operate automatically because equipped perception, planning, and control modules operate cooperatively \cite{Gri20, Yur20, Tam22}. The most common perception components used in autonomous vehicles include cameras and radar/lidar devices; cameras are combined with DNN to recognize relevant objects, and radars/lidars are mainly used for distance measurement \cite{Yur20, YLi20}. Because of limitations related to sensor cost and size, current Active Driving Assistance Systems (ADASs) primarily rely on camera-based perception modules with supplementary radars \cite{Con20}.

To understand complex driving scenes, multi-task DNN (MTDNN) models that output multiple predictions simultaneously are often applied in autonomous vehicles to reduce inference time and device power consumption. In \cite{Tei18}, street classification, vehicle detection, and road segmentation problems were solved using a single MultiNet model. In \cite{Piz19}, the researchers trained an MTDNN to detect drivable areas and road classes for vehicle navigation. DLT-Net, presented in \cite{Qia20}, is a unified neural network for the simultaneous detection of drivable areas, lane lines, and traffic objects. The network localizes the vehicle when a high-definition (HD) map is unavailable. The context tensors between subtask decoders in DLT-Net share mutual features learned from different tasks. A lightweight multi-task semantic attention network was proposed in \cite{Lai21} to achieve simultaneous object detection and semantic segmentation; this network boosts detection performance and reduces computational costs through the use of a semantic attention module. YOLOP \cite{Don21} is a panoptic driving perception network that simultaneously performs traffic object detection, drivable area segmentation, and lane detection on an NVIDIA TITAN XP GPU at a speed of 41 FPS (frames per second). In the commercially available TESLA  Autopilot system \cite{tesla},  images from cameras with different viewpoints are entered into separate MTDNNs that perform driving scene semantic segmentation, monocular depth estimation, and object detection tasks. The outputs of these MTDNNs are further fused in bird’s-eye-view (BEV) networks to directly output a reconstructed aerial-view map of traffic objects, static infrastructure, and the road itself.

In a modular self-driving system, the environmental perception results can be sent to an optimization-based model predictive control (MPC) planner to generate spatiotemporal curves over a time horizon.  The  system then reactively selects optimal solutions over a short interval as control inputs to minimize the gap between target and current states \cite{Pad16}. These MPC models can be realized with various methods [e.g., active set, augmented Lagrangian, interior point, or sequential quadratic programming (SQP)] \cite{Lim21,Gut17} and are promising for vehicle optimal control problems. In \cite{Tur13}, a linear MPC control model was proposed that addresses vehicle lane-keeping and obstacle avoidance problems by using lateral automation. In \cite{Kat13}, an MPC control scheme combining longitudinal and lateral dynamics was designed for following velocity trajectories. Ref. \cite{SLi15}  proposed a scale reduction method for reducing the online computational efforts of MPC controllers, and they applied it to longitudinal vehicle automation, achieving an average computational time of approximately 4 ms. In \cite{Gut17}, a linear time-varying MPC scheme was proposed for lateral automobile trajectory optimization. The cycle time for the optimized trajectory to be communicated to the feedback controller was 10 ms. In addition, \cite{Lim22} investigated automatic weight determination for car-following control, and the corresponding linear MPC algorithm was implemented using CVXGEN \cite{Mat12}, which solves the relevant problem within 1 ms.

The constrained iterative linear quadratic regulator (CILQR) method was proposed to solve online trajectory optimization problems with nonlinear system dynamics and general constraints \cite{Che17,Che19}. The CILQR algorithm constructed on the basis of differential dynamic programming (DDP) \cite{Jac70} is also an MPC method. The computational load of the well-established SQP solver is higher than that of DDP \cite{Ma22}. Thus, the CILQR solver outperforms the standard SQP approach in terms of computational efficiency; compared with the CILQR solver, the SQP approach requires a computation time that is 40.4 times longer per iteration \cite{Che19}. However, previous CILQR-relates studies \cite{Che17,Che19, Ma22,Pan20} have focused on nonlinear Cartesian-frame motion planning. Alternatively, planning within the Fren{\'e}t-frame can reduce problem dimensions because it enables vehicle dynamics to be solved in tangential and normal directions separately with the aid of road reference line  \cite{Wer10}; furthermore, the corresponding linear dynamic equations \cite{Lim22,Lee19} do not have adverse effects when high-order Taylor expansion coefficients are truncated in the CILQR framework [cf. Section II]. These considerations motivated us to use linear CILQR planners to control automated vehicles.

\begin{figure}[t]
\centerline{\includegraphics[scale=0.3]{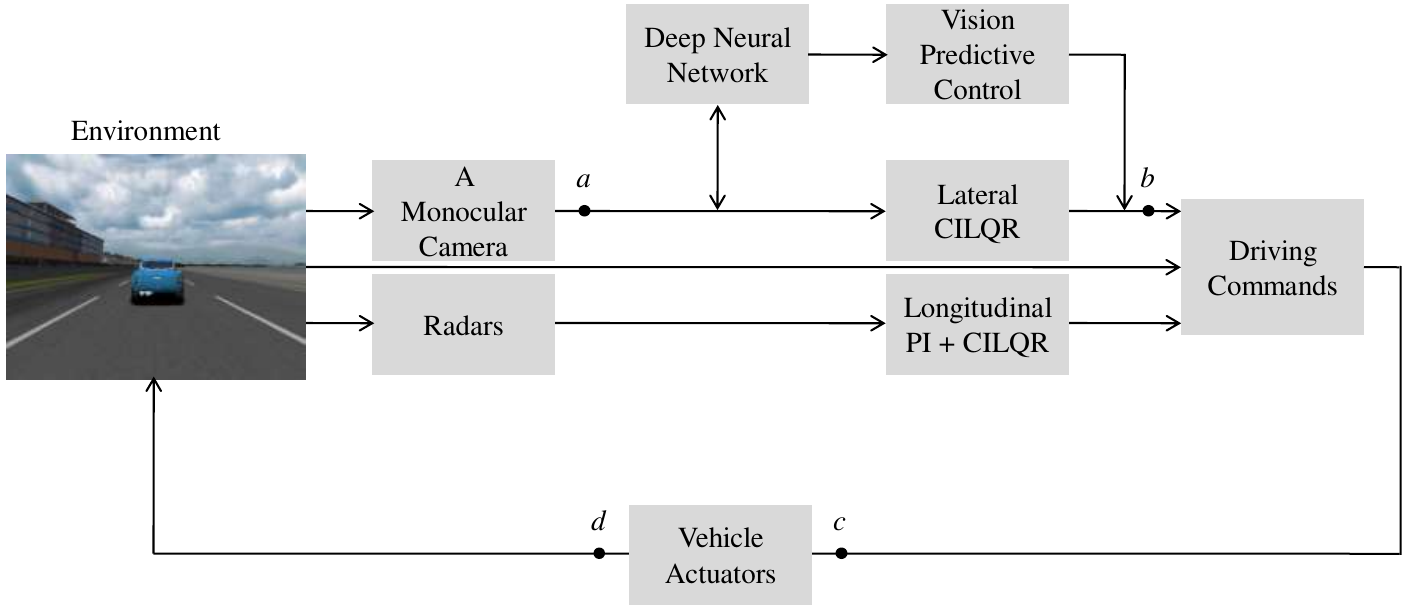}}
\caption{Proposed vision-based automated driving framework. The system comprises the following modules: a multi-task DNN for perceiving surroundings, vision predictive control and CILQR controllers for vehicle motion planning and adherence to driving commands (steering, acceleration, and braking), and a  PI  controller combined with the longitudinal CILQR algorithm for velocity tracking.  These modules receive input data from a monocular camera and a few inexpensive radars and  operate collaboratively to operate the automated vehicle. The DNN, vision predictive control, and lateral and longitudinal CILQR algorithms are run efficiently every 24.52, 15.56, and 0.58 and 0.65 ms, respectively. In our simulation, the end‐to‐end latency  from the camera output to the lateral controller output ($T_{a \to b}  \equiv T_{lat} $) is longer than the actuator latency ($T_{c\to d}  \equiv T_{act}= 6.66 $ ms).}
\end{figure}

We proposed an MTDNN in \cite{Lee21} to directly perceive ego vehicle’s heading angle ($\theta $) and  distance  from the lane centerline ($\Delta $) for autonomous driving. The vision-based  MTDNN model in \cite{Lee21} essentially provides the information necessary for ego car navigation within Fren{\'e}t coordinates without the need for HD maps. Nevertheless, this end-to-end autonomous driving approach performs poorly in environments that are not shown during the training phase \cite{Yur20}. In \cite{Lee21a}, we  proposed  an improved control algorithm based on a multi-task  UNet architecture (MTUNet) that comprises lane line segmentation and pose estimation subnets. A Stanley controller \cite{Thr06}  was then designed to control the lateral automation of an automobile. The Stanley controller takes  $\theta $ and  $\Delta $ yielding from the network as it's input for lane-centering \cite{Lee22}. The improved algorithm outperforms the model in \cite{Lee21} and has comparable performance to a multi-task-learning reinforcement-learning (MTL-RL) model \cite{Li19}, which integrates RL and deep-learning algorithms for autonomous driving. However, our algorithms presented in \cite{Lee21a}  have a variety of problems as follows. 1) Vehicle dynamic models are not considered in the Stanley controller, and the model has poor performance for lanes with  rapid curvature changes \cite{Liu21}. 2) The proposed self-driving system does not consider road curvature, resulting in poor vehicle  control on curvy roads \cite{Lu21}. 3)  The corresponding DNN perception network lacks object detection capability, which is a core task in automated driving. 4) The DNN input has  high dimensional resolution of 400 $\times$ 400, which results in long training and inference times.

\begin{figure}[!t]
\centerline{\includegraphics[scale=0.3]{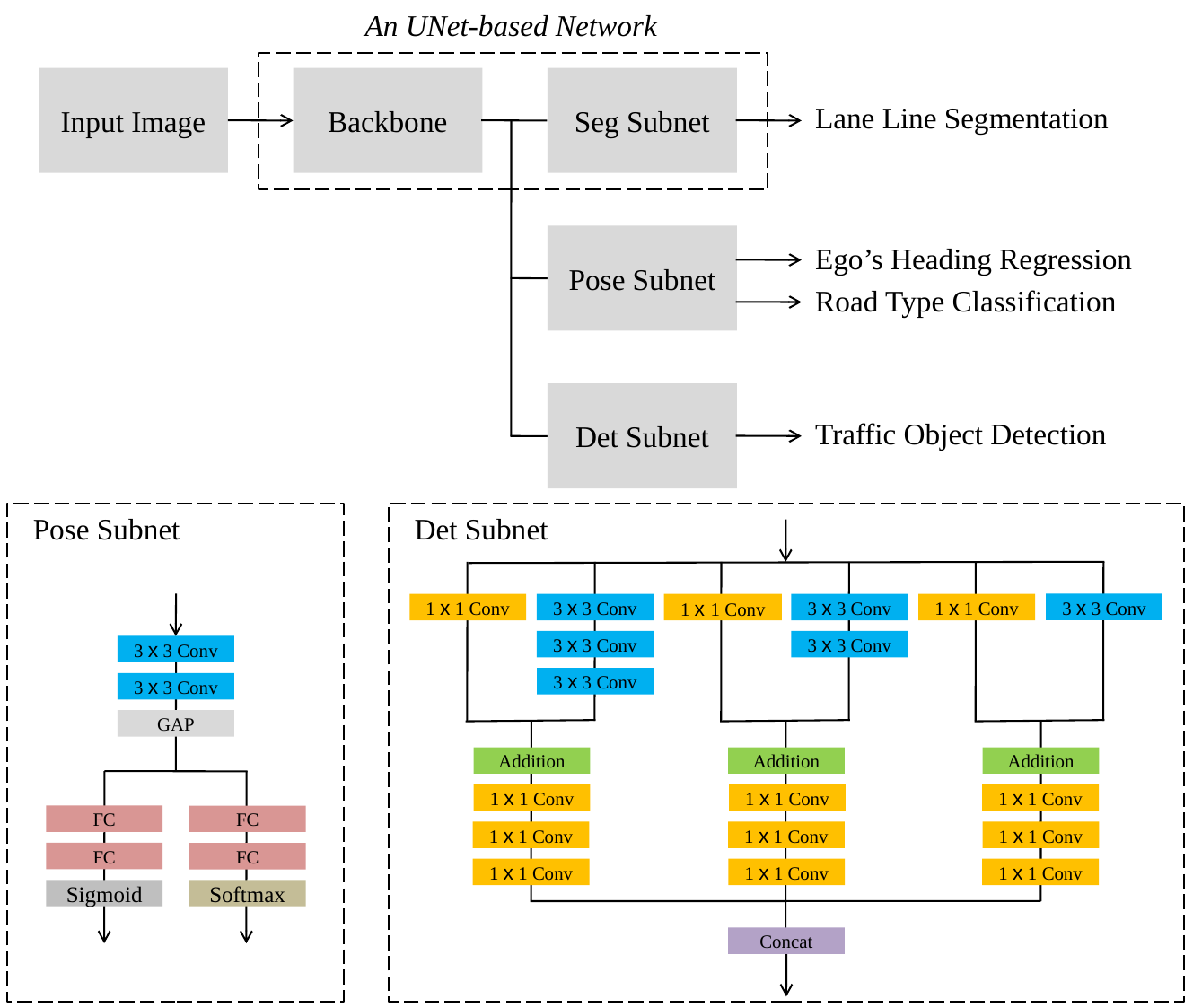}}
\caption{Overview of proposed MTUNet architecture. The input RGB image of size  228 $\times$ 228 is fed into the model, which then performs lane line segmentation, ego vehicle's pose estimation,  and traffic object detection at the same time. The backbone-seg-subnet is an UNet-based network; three variants of UNet (UNet$\_$2$\times$ \cite{Lee21a}, UNet$\_$1$\times$ \cite{Nab20}, and MResUNet \cite{Nab20}) are compared in this work. The ReLU  activation functions in  pose and det subnets are not shown for simplicity.}
\end{figure}

To address these shortcomings, this paper proposes a new system for real-time automated driving based on the developments described in \cite{Lee21a, Lee22}. First, a YOLOv4  detector \cite{Ale20} is added to the MTUNet for object detection. Second, the inference speed of  MTUNet was increased by reducing the input size without sacrificing network performance.  Third, a vision predictive control (VPC) algorithm is proposed for reducing the steering command delay by enabling steer correction at a look-ahead point by applying road curvature information. The VPC algorithm can also be combined with the lateral CILQR algorithm (denoted VPC-CILQR) to rapidly perform motion planning and automobile control. As shown in Fig. 1, the vehicle actuation latency ($T_{act}$) was shorter than the steering command latency ($T_{lat}$) in our simulation. This delay may also be present in automated vehicles \cite{JMG20} or autonomous racing systems \cite{Bet23} and may induce instability in the system being controlled. Equipping the vehicle with low-level computers could further increase this steering command lag. Therefore, compensating algorithms such as VPC are key to cost-efficient automated vehicle systems.

In general, the research method of this paper is similar to those in  \cite{Zha19, Xu20}, which have also presented self-driving systems based on lane detection results. In \cite{Zha19}. an optimal LQR scheme with the sliding-mode approach was proposed to implement preview path tracking control for intelligent electric vehicles with optimal torque distribution between their motors. In \cite{Xu20}, a safeguard-protected preview path tracking control algorithm was presented. The proposed preview control strategy comprises feedback and feedforward controllers for stabilizing tracking errors and preview control, respectively. The proposed controller was implemented and validated on open roads and Mcity, an automated vehicle platform. The tested vehicle was equipped with a commercial Mobileye module to detect lane markings.

The main goal of this work was to design a computationally efficient automated driving system for real-time lane-keeping and car-following. The contributions of this paper are as follows:
\begin{enumerate}
  \item The proposed MTDNN scheme can execute simultaneous driving perception tasks at a speed of 40 FPS. The main difference between this scheme and previous MTDNN schemes is that the post-processing methods provide crucial parameters (lateral offset, road curvature, and heading angle) that improve local vehicular navigation.  
  \item The VPC-CILQR controller comprising the VPC algorithm and lateral CILQR solver is proposed to improve driverless vehicle path tracking. The method has a low online computational burden and can respond to steering commands in accordance with the concept of look-ahead distance.
  \item We propose a vision-based framework comprising the aforementioned MTDNN scheme and CILQR-based controllers for operating an autonomous vehicle; the effectiveness of the proposed framework was demonstrated in challenging simulation environments without maps.
\end{enumerate}

The remainder of this paper is organized as follows: the research methodology is presented in Section II, the experimental setup is described in Section III, and the results are presented and discussed in Section IV. Section V concludes this paper.

\section{Methodology}
The following section introduces each part of the proposed self-driving system. As depicted in Fig. 1, our system comprises several modules. The DNN is an MTUNet that can solve multiple perception problems simultaneously. The CILQR controllers receive data from the DNN and radars to compute driving commands for lateral and longitudinal motion planning. In the lateral direction, the lane line detection results from the DNN are input to the VPC module to compute steering angle corrections at a certain distance in front of the ego car. These corrections are then sent to the CILQR solver to predict a steering angle for the    lane-keeping task. This two-step algorithm is denoted VPC-CILQR throughout the article. The other CILQR controller handles the car-following task in the longitudinal direction.

\subsection{MTUNet network}
As indicated in Fig. 2, the proposed MTDNN is a neural network with an MTUNet architecture featuring a common backbone encoder and three subnets for completing multiple tasks at the same time. The following sections describe each part.

\subsubsection{Backbone and segmentation subnet}
The shared backbone and segmentation (seg) subnet employ encoder-decoder UNet-based networks for  pixel-level lane line classification task. Two classical  UNets (UNet$\_$2$\times$ \cite{Lee21a} and UNet$\_$1$\times$ \cite{Nab20})  and one enhanced version (MultiResUNet \cite{Nab20}, denoted as MResUNet throughout the paper)   were used to investigate the effects of model size and complexity on task performance. For UNet$\_$2$\times$ and UNet$\_$1$\times$, each repeated block includes two convolutional (Conv) layers, and the first UNet has twice as many filters as the second. For MResUNet, each modified block consists of three 3 $\times$ 3 Conv layers and one 1 $\times$ 1 Conv layer. Table I summarizes the  filter number and related kernel size of the Conv layers used in these models. The resulting total number of parameters of UNet$\_$2$\times$/UNet$\_$1$\times$/MResUNet is 31.04/7.77/7.26 M, and the corresponding total number of multiply accumulate operations (MACs) is 38.91/9.76/12.67 G. All 3 $\times$ 3 Conv layers are padded with one pixel to preserve the spatial resolution after the convolution operations are applied \cite{Sim15}. This setting reduces the network input size from 400 $\times$ 400 to 228 $\times$ 228 but preserves model performance and increases inference speed compared with the models in our previous work (the experimental results are presented in Section IV) \cite{Lee21a}. That network used unpadded 3 $\times$ 3 Conv layers \cite{Ron15}, and zero padding was therefore applied to the input to equalize the input--output resolutions \cite{Bru18}.  In the training phase, the weighted cross-entropy loss is adopted to deal with the lane detection sample imbalance problem  \cite{Xie15, Zou20} and is represented as
\begin{equation} 
\begin{split}
L_S  = & - \frac{N^{-}}{N^{+}+N^{-}}\sum\limits_{\tilde y=1} \log \left(\sigma(y) \right) \\
 & - \frac{N^{+}}{N^{+}+N^{-}} \sum\limits_{\tilde y=0} \log \left(1-\sigma(y) \right) ,
\end{split}
\end{equation}
where  $N^{+}$ and $N^{-}$ are the numbers of foreground and background samples in a batch of images, respectively; $y$ is a predicted score; $\tilde y$ is the corresponding label; and $\sigma$ is the sigmoid function.

\subsubsection{Pose subnet} 
This subnet is mainly responsible for whole-image angle regression and road type classification problems, where the road involves three categories (left turn, straight, and right turn) designed to prevent the angle estimation from mode collapsing \cite{Lee21a,Cui19}. The network architecture of the pose subnet is presented in Fig. 2; the pose subnet takes the fourth Conv-block output feature maps of the backbone as its input. Subsequently, the input maps are fed into shared parts including two consecutive Conv layers and one global average pooling (GAP) layer to extract general features. Lastly, the resulting vectors are passed  separately through two fully connected (FC) layers before being mapped into a sigmoid/softmax activation layer for the regression/classification task. Table II summarizes the number of filters  and output units of the corresponding Conv and FC layers, respectively. The expression   MTUNet$\_$2$\times$/MTUNet$\_$1$\times$/MTMResUNet in Table II represents a multi-task UNet scheme in which subnets are built on the UNet$\_$2$\times$/UNet$\_$1$\times$/MResUNet model throughout the article. The pose task loss function, including L2 regression loss ($L_R$) and cross-entropy loss ($L_C$), is employed for network training; this function is represented as follows:
\begin{subequations}
\begin{equation}
L_R  = \frac{1}{{2B}}\sum\limits_{i = 1}^B {\left| {\sigma(\tilde{\theta}_i)- \sigma(\theta_i) } \right|^2 },
\end{equation}
\begin{equation}
L_C  =  -\frac{1}{B}\sum\limits_{i = 1}^B \sum\limits_{j = 1}^3 \tilde p_{ij} \log(p_{ij}),
\end{equation}
\end{subequations}
where $\tilde \theta $ and $\theta$ are the ground truth and estimated value, respectively; $B$ is the input batch size; and $\tilde p$ and $p$ are true and softmax estimation values, respectively.

\subsubsection{Detection subnet}
The detection (det) subnet takes advantage of a simplified YOLOv4 detector \cite{Ale20} for real-time traffic object (leading car) detection. This fully-convolutional subnet that has three branches for multi-scale detection takes the output feature maps of the backbone as its input, as illustrated in Fig. 2. The initial part of each branch is composed of single or consecutive 3 $\times$ 3 filters for extracting contextual information at different scales  \cite{Nab20}, and a  shortcut connection with one 1 $\times$ 1 filter from the input layer for residual mapping. The top of the addition layer contains sequential 1 $\times$ 1 filters for reducing the number of channels. The resulting feature maps of each branch have six channels (five for bounding box offset and confidence score predictions, and one for class probability estimation) with a size of $K$ = 15 $\times$ 15 to divide the input image into $K$ grids. In this article, we select $M$ = 3 anchor boxes, which are then shared between three branches according to the context size. Ultimately, spatial features from three detecting scales are concatenated together and sent to the output layer. Table  III presents the design of detection subnet of MTUNets. The overall  loss function for training comprises objectness ($L_{O}$), classification ($L_{CL}$), and complete intersection over union (CIoU) losses ($L_{CI}$) \cite{Zhe20, Jun22}; these losses are constructed as follows:
\begin{subequations}
\begin{align}
\begin{split}
 L_{O}  = & - \sum\limits_{i = 1}^{K \times M} { {I_{i}^{o} \left[ {\tilde Q_i \log \left( {Q_i } \right)} \right.} }  
 \left. { + \left( {1 - \tilde Q_i } \right)\log \left( {1 - Q_i } \right)} \right] \\ &
  - { {\lambda_{n} I_{i}^{n} \left[ {\tilde Q_i \log \left( {Q_i } \right)} \right.} }  
 \left. { + \left( {1 - \tilde Q_i } \right)\log \left( {1 - Q_i } \right)} \right], \\ 
\end{split}
\end{align} 
\begin{align}
\begin{split}
 L_{CL}   = &- \sum\limits_{i = 1}^{K \times M} {    I_{i}^{o} \sum\limits_{c \in classes}  {{\tilde p_i \left( c \right)\log \left( {p_i \left( c \right)} \right)} } }  \\ 
  & + \left( {1 - \tilde p_i \left( c \right)} \right)\log \left( {1 - p_i \left( c \right)} \right) ,  
\end{split}
\end{align} 
\begin{equation}
L_{CI}  = 1 - IoU + \frac{{E ^2 \left( {{\bf \tilde o},{\bf o}} \right)}}{{\beta ^2 }} + \alpha \gamma,
\end{equation}
\end{subequations}
where $I_{i}^{o/n}$ = 1/0 or 0/1 indicates that the $i$-th predicted bounding box does or does not contain an object, respectively; $\tilde Q_i$/$Q_i$ and $\tilde p_i$/$p_i$ are the true/estimated objectness and class scores corresponding to each box, respectively; and $\lambda_{n}$ is a hyperparameter intended for balancing positive and negative samples. With regard to CIoU loss, ${\bf \tilde o}$ and ${\bf o} $ are the central points of the prediction ($B_p$) and ground truth ($B_{gt}$) boxes, respectively; $E$ is the related Euclidean distance; $\beta $ is the diagonal distance of the smallest enclosing box covering $B_p$ and $B_{gt}$; $\alpha$ is a  tradeoff hyperparameter; and $\gamma$ is used to  measure  aspect ratio  consistency \cite{Zhe20}.

\begin{table}[!t]
\caption{Conv layers used in the UNet-based networks}
\begin{center}
\begin{tabular}{l|c|c|c}
\hline
\multirow{2}{*}{Conv-block}  &\multirow{2}{*}{UNet$\_$2$\times$ \cite{Lee21a}}   &\multirow{2}{*}{UNet$\_$1$\times$ \cite{Nab20}}  &\multirow{2}{*}{MResUNet \cite{Nab20}}   \\  
 & & &   \\  \hline
\multirow{2}{*}{Block$1^{a}$/9}  & 64 (3 $\times$ 3)$^{b}$  & 32 (3 $\times$ 3)  & 8 (3 $\times$ 3), 17 (3 $\times$ 3),   \\ 
   &  64 (3 $\times$ 3)  & 32 (3 $\times$ 3)   &  26 (3 $\times$ 3), 51 (1 $\times$ 1)  \\ 
\multirow{2}{*}{Block2/8}  & 128 (3 $\times$ 3)  & 64 (3 $\times$ 3)  & 17 (3 $\times$ 3), 35 (3 $\times$ 3),   \\ 
   &  128 (3 $\times$ 3)  & 64 (3 $\times$ 3)   &  53 (3 $\times$ 3), 105 (1 $\times$ 1)  \\ 
\multirow{2}{*}{Block3/7}  & 256 (3 $\times$ 3)  & 128 (3 $\times$ 3)  & 35 (3 $\times$ 3), 71 (3 $\times$ 3),   \\ 
   &  256 (3 $\times$ 3)  & 128 (3 $\times$ 3)   &  106 (3 $\times$ 3), 212 (1 $\times$ 1)  \\ 
\multirow{2}{*}{Block4/6}  & 512 (3 $\times$ 3)  & 256 (3 $\times$ 3)  & 71 (3 $\times$ 3), 142 (3 $\times$ 3),   \\ 
   &  512 (3 $\times$ 3)  & 256 (3 $\times$ 3)   &  213 (3 $\times$ 3), 426 (1 $\times$ 1)  \\ 
\multirow{2}{*}{Block5}  & 1024 (3 $\times$ 3)  & 512 (3 $\times$ 3)  & 142 (3 $\times$ 3), 284 (3 $\times$ 3),   \\ 
   &  1024 (3 $\times$ 3)  & 512 (3 $\times$ 3)   &  427 (3 $\times$ 3), 853 (1 $\times$ 1)  \\ \hline

\multicolumn{4}{l}{$^{a}$\scriptsize{Block1 of UNet$\_$2$\times$/UNet$\_$1$\times$ only contains one 3 $\times$ 3 Conv layer}}  \\ 
\multicolumn{4}{l}{$^{b}$\scriptsize{The notation $n$ ($k$ $\times$ $k$) represents a Conv layer with $n$ filters of kernel size $k$ $\times$ $k$  }}
\end{tabular}
\end{center}
\end{table}

\begin{table}[!t]
\caption{Conv and FC layers used in the pose subnet of various MTUNets}
\begin{center}
\begin{tabular}{l|c|c}
\hline
\multirow{2}{*}{Layer}  & \multirow{2}{*}{MTUNet$\_$2$\times$}   & MTUNet$\_$1$\times$ \\  
 &  & MTMResUNet  \\  \hline

 Conv1  &  1024 (3 $\times$ 3)  & 512 (3 $\times$ 3)     \\ 
 Conv2  &  1024 (3 $\times$ 3)  & 512 (3 $\times$ 3)     \\ 
 FC1-2  &  256, 256  & 256, 256   \\ 
 FC3-4  &  1, 3  & 1, 3   \\ \hline
\end{tabular}
\end{center}
\end{table}

\begin{table}[!t]
\caption{Conv layers used in the detection subnet  of various MTUNets}
\begin{center}
\begin{tabular}{l|c|c}
\hline
\multirow{2}{*}{Layer}   & \multirow{2}{*}{MTUNet$\_$2$\times$}   & MTUNet$\_$1$\times$ \\  
 &  & MTMResUNet  \\  \hline
 Conv1-6  &  512 (3 $\times$ 3)  & 384 (3 $\times$ 3)   \\ 
 Conv7-9  &  512 (1 $\times$ 1)  & 384 (1 $\times$ 1)     \\ 
 Conv10-12  &  256 (1 $\times$ 1)  & 256 (1 $\times$ 1)     \\ 
 Conv13-15  &  256 (1 $\times$ 1)  & 256 (1 $\times$ 1)     \\ 
 Conv16-18  &  6 (1 $\times$ 1)  & 6 (1 $\times$ 1)     \\ \hline
\end{tabular}
\end{center}
\end{table}

\subsection{The CILQR algorithm}

This section first briefly describes the concept behind  CILQR and related approaches based on  \cite{Che17,Che19,Tas12,Tas14}; it then presents the lateral/longitudinal CILQR control algorithm that takes the MTUNet inference and radar data as its inputs to yield driving decisions using linear dynamics.

\subsubsection{Problem formulation}
Provided  a sequence of states ${\bf X} \equiv \left\{ {{\bf x}_0 ,{\bf x}_1 ,...,{\bf x}_N } \right\}$ and the corresponding control sequence ${\bf U} \equiv \left\{ {{\bf u}_0 ,{\bf u}_1 ,...,{\bf u}_{N - 1} } \right\}$  are within the preview horizon $N$,  the  system's discrete-time dynamics ${\bf f}$ are satisfied, with
\begin{equation}
{\bf x}_{i + 1}  = {\bf f}\left( {{\bf x}_i ,{\bf u}_i } \right)
\end{equation}
from time $i$ to $i+1$. The total cost denoted by $\mathcal{J}$, including running costs $\mathcal{P}$ and the final cost $\mathcal{P}_f$, is presented as follows:
\begin{equation}
\mathcal{J}\left( {{\bf x}_0 ,{\bf U}} \right) = \sum\limits_{i = 0}^{N - 1} {\mathcal{P}\left( {{\bf x}_i ,{\bf u}_i } \right) + \mathcal{P}_f \left( {{\bf x}_N } \right)}. 
\end{equation}
The  optimal  control sequence is then written as 
\begin{equation}
{\bf U}^* \left( {{\bf x}^* } \right) \equiv \mathop {\arg \min }\limits_{\bf U} \mathcal{J}\left( {{\bf x}_0 ,{\bf U}} \right)
\end{equation}
with an optimal trajectory ${{\bf x}^* }$. The partial sum of $\mathcal{J}$ from any time step $t$
to $N$ is represented as
\begin{equation}
\mathcal{J}_t \left( {{\bf x},{\bf U}_t } \right) = \sum\limits_{i = t}^{N - 1} {\mathcal{P}\left( {{\bf x}_i ,{\bf u}_i } \right) + \mathcal{P}_f \left( {{\bf x}_N } \right)}, 
\end{equation}
and the optimal value function $\mathcal{V}$ at time $t$ starting at ${\bf x}$ takes the form
\begin{equation}
\mathcal{V}_t \left( {\bf x} \right) \equiv \mathop {\arg \min }\limits_{{\bf U}_t } \mathcal{J}_t \left( {{\bf x},{\bf U}_t } \right)
\end{equation}
with  the final time step value function  $\mathcal{V}_N \left( {\bf x} \right) \equiv \mathcal{P}_f \left( {{\bf x}_N } \right)$.  

In practice, the  final step value function $\mathcal{V}_N\left( {\bf x} \right)$ is obtained by executing a forward pass using the current control sequence. Subsequently,  local control signal minimizations are performed in the proceeding backward pass using the following Bellman equation:
\begin{equation}
\mathcal{V}_i \left( {\bf x} \right) = \mathop {\min }\limits_{\bf u} \left[ {\mathcal{P}\left( {{\bf x},{\bf u}} \right) + \mathcal{V}_{i + 1} \left( {{\bf f}\left( {{\bf x},{\bf u}} \right)} \right)} \right].
\end{equation}
To compute the optimal  trajectory,  the perturbed function around the $i$-th state-control pair in Eq. (9) is used; this function is written as follows: 
\begin{align}
\begin{split}
\mathcal{O}\left( {\delta {\bf x},\delta {\bf u}} \right) =& \mathcal{P}_i \left( {{\bf x} + \delta {\bf x},{\bf u} + \delta {\bf u}} \right) - \mathcal{P}_i \left( {{\bf x},{\bf u}} \right) \\ &+ \mathcal{V}_{i + 1} \left( {{\bf f}\left( {{\bf x} + \delta {\bf x},{\bf u} + \delta {\bf u}} \right)} \right) - \mathcal{V}_{i + 1} \left( {{\bf f}\left( {{\bf x},{\bf u}} \right)} \right).
\end{split}
\end{align} 
This equation can  be  approximated to a quadratic function by employing a second-order Taylor expansion with the  following coefficients:
\begin{subequations}
\begin{equation}
\mathcal{O}_{\bf x}  = \mathcal{P}_{\bf x}  + {\bf f}_{\bf x}^{\rm T} \mathcal{V}_{\bf x}, 
\end{equation}
\begin{equation}
\mathcal{O}_{\bf u}  = \mathcal{P}_{\bf u}  + {\bf f}_{\bf u}^{\rm T} \mathcal{V}_{\bf x},
\end{equation}
\begin{equation}
\mathcal{O}_{{\bf xx}}  = \mathcal{P}_{{\bf xx}}  + {\bf f}_{\bf x}^{\rm T} \mathcal{V}_{{\bf xx}} {\bf f}_{\bf x}  + \mathcal{V}_{\bf x}   {\bf f}_{{\bf xx}},
\end{equation}
\begin{equation}
\mathcal{O}_{{\bf ux}}  = \mathcal{P}_{{\bf ux}}  + {\bf f}_{\bf u}^{\rm T} \mathcal{V}_{{\bf xx}} {\bf f}_{\bf x}  + \mathcal{V}_{\bf x}   {\bf f}_{{\bf ux}}, 
\end{equation}
\begin{equation}
\mathcal{O}_{{\bf uu}}  = \mathcal{P}_{{\bf uu}}  + {\bf f}_{\bf u}^{\rm T} \mathcal{V}_{{\bf xx}} {\bf f}_{\bf u}  + \mathcal{V}_{\bf x}   {\bf f}_{{\bf uu}}. 
\end{equation}
\end{subequations}
The second-order  coefficients of the system dynamics (${\bf f}_{\bf xx}$, ${\bf f}_{\bf ux}$, and ${\bf f}_{\bf uu}$) are omitted  to reduce computational effort \cite{Ma22,Tas12}. The values of these  coefficients are zero for linear systems [e.g., Eq. (19) and Eq. (25)], leading to fast convergence in trajectory optimization. 

The optimal control signal modification can  be obtained by minimizing the quadratic
 $\mathcal{O}\left( {\delta {\bf x},\delta {\bf u}} \right)$:
\begin{equation}
\delta {\bf u}^*  = \mathop {\arg \min }\limits_{\delta {\bf u}} \mathcal{O}\left( {\delta {\bf x},\delta {\bf u}} \right) = {\bf k} + {\bf K}\delta {\bf x},
\end{equation}
where 
\begin{subequations}
\begin{equation}
{\bf k} =  - \mathcal{O}_{{\bf uu}}^{ - 1} \mathcal{O}_{\bf u},
\end{equation}
\begin{equation}
{\bf K} =  - \mathcal{O}_{{\bf uu}}^{ - 1} \mathcal{O}_{{\bf ux}}  
\end{equation}
\end{subequations}
are optimal control gains. If the optimal control indicated in Eq. (12) is plugged into  the approximated $\mathcal{O}\left( {\delta {\bf x},\delta {\bf u}} \right)$  to recover the quadratic value function, the corresponding coefficients can  be obtained \cite{Pla17}:
\begin{subequations}
\begin{equation}
\mathcal{V}_{\bf x}  = \mathcal{O}_{\bf x}  - {\bf K}^{\rm T} \mathcal{O}_{{\bf uu}} {\bf k},
\end{equation}
\begin{equation}
\mathcal{V}_{{\bf xx}}  = \mathcal{O}_{{\bf xx}}  - {\bf K}^{\rm T} \mathcal{O}_{{\bf uu}} {\bf K}.
\end{equation}
\end{subequations}
Control gains at each state (${\bf k}_i$, ${\bf K}_i$) can then be estimated by recursively computing Eqs. (11), (13), and (14) in a backward process. Finally, the modified  control and state sequences can be evaluated through a renewed forward pass: 
\begin{subequations}
\begin{equation}
{\bf \hat u}_i  = {\bf u}_i  +\lambda {\bf k}_i  + {\bf K}_i \left( {{\bf \hat x}_i  - {\bf x}_i } \right),
\end{equation}
\begin{equation}
{\bf \hat x}_{i + 1}  = {\bf f}\left( {{\bf \hat x}_i ,{\bf \hat u}_i } \right),
\end{equation}
\end{subequations}
where  ${\bf \hat x}_0  = {\bf x}_0 $. Here $\lambda$ is the backtracking parameter for line search; it is set to 1 in the beginning and  designed to be reduced gradually in  the   forward-backward propagation loops  until convergence is reached.

If the system has the constraint 
\begin{equation}
 \mathcal{C}\left( {x,u} \right) < 0,
\end{equation}
which can be shaped using an exponential barrier function \cite{Che17,Pan20}
\begin{equation}
\mathcal{B}\left( {\mathcal{C}\left( {x,u} \right)} \right) = q_1 \exp \left( {q_2 \mathcal{C}\left( {x,u} \right)} \right)
\end{equation}
or a logarithmic barrier function \cite{Che19}, then 
\begin{equation}
\mathcal{B}\left( {\mathcal{C}\left( {x,u} \right)} \right) =  - \frac{1}{t}\log \left( { - \mathcal{C}\left( {x,u} \right)} \right),
\end{equation}
where $q_1$, $q_2$, and $t > 0$ are parameters. The barrier function can be added to the cost function as a penalty. Eq. (18) converges toward the ideal indicator function as $t$ increases iteratively.

\subsubsection{Lateral CILQR controller}
The lateral vehicle dynamic model \cite{Lee19} is employed for steering control. The state variable and control input are defined as $
{\bf x} = \left[ {\begin{array}{*{20}c}
   \Delta  & {\dot \Delta } & \theta  & {\dot \theta }  \\
\end{array}} \right]^{\rm T} 
$ and ${\bf u} = \left[  \delta  \right]$, respectively, where $\Delta$ is the lateral offset, $\theta $ is the angle between the ego vehicle’s heading and the tangent of the road, and $\delta$ is the steering angle. As described in  our previous work \cite{Lee21a, Lee22},  $\theta $ and $\Delta$ can be obtained from MTUNets and related post-processing methods, and it is assumed that $\dot \Delta  = \dot \theta  = 0$. The corresponding discrete-time dynamic model is written as follows: 
\begin{equation}
{\bf x}_{t + 1}  \equiv {\bf f}\left( {{\bf x}_t ,{\bf u}_t } \right) = {\bf Ax}_t  + {\bf Bu}_t,
\end{equation}
where
\[
{\bf A} = \left[ {\begin{array}{*{20}c}
{\alpha _{11} } & {\alpha _{12} } & 0 & 0 \\
0 & {\alpha _{22} } & {\alpha _{23} } & {\alpha _{24} } \\
0 & 0 & {\alpha _{33} } & {\alpha _{34} } \\
0 & {\alpha _{42} } & {\alpha _{43} } & {\alpha _{44} } \\
\end{array}} \right],\quad{\bf B} = \left[ {\begin{array}{*{20}c}
0 \\
{\beta _1 } \\
0 \\
{\beta _2 } \\
\end{array}} \right],
\]
with coefficients
\[
\begin{array}{l}
\alpha _{11} = \alpha _{33} = 1, \quad\alpha _{12} = \alpha _{34} = dt, \\
\alpha _{22} = {1 - \frac{{2\left( {C_{\alpha f} + C_{\alpha r} } \right)dt}}{{mv }}},\quad\alpha _{23} = {\frac{{2\left( {C_{\alpha f} + C_{\alpha r} } \right)dt}}{m}}, \\
\alpha _{24} = {\frac{{2\left( { - C_{\alpha f} l_f + C_{\alpha r} l_r } \right)dt}}{{mv }}},\quad\alpha _{42} = {\frac{{2\left( {C_{\alpha f} l_f - C_{\alpha r} l_r } \right)dt}}{{I_z v }}} , \\
\alpha _{43} = {\frac{{2\left( {C_{\alpha f} l_f - C_{\alpha r} l_r } \right)dt}}{{I_z }}},\quad\alpha _{44} = {1 - \frac{{2\left( {C_{\alpha f} l_f^2 - C_{\alpha r} l_r^2 } \right)dt}}{{I_z v }}}, \\
\beta _1 = {\frac{{2C_{\alpha f} dt}}{m}} ,\quad\beta _2 = {\frac{{2C_{\alpha f} l_f dt}}{{I_z }}}. \\
\end{array}
\]
Here, $v$ is the ego vehicle's current speed along the heading direction and $dt$ is the sampling time. The model parameters for the experiments are as follows: vehicle mass $m$ = 1150 kg, cornering stiffness ${C_{\alpha f} }$ = 80\thinspace000 N/rad, ${C_{\alpha r} }$ = 80\thinspace000 N/rad, center of gravity point $l_f$ = 1.27 m, $l_r$ = 1.37 m, and moment of inertia $I_z$ = 2000 kgm$^{2}$.

The objective function ($\mathcal{J}$) containing the iterative linear quadratic regulator ($\mathcal{J}_{ILQR}$),  barrier ($\mathcal{J}_{b}$), and  end state cost ($\mathcal{J}_{f}$) terms can be represented as
\begin{subequations}
\begin{equation}
\mathcal{J} = \mathcal{J}_{ILQR}  + \mathcal{J}_{b}+ \mathcal{J}_{f}, 
\end{equation}
\begin{equation}
\mathcal{J}_{ILQR}  = \sum\limits_{i = 0}^{N - 1} {\left( {{\bf x}_i  - {\bf x}_{r} } \right)^{\rm T} {\bf Q}\left( {{\bf x}_i  - {\bf x}_{r} } \right) + {\bf u}_i^{\rm T} {\bf Ru}_i }, 
\end{equation}
\begin{equation}
\mathcal{J}_{b}  = \sum\limits_{i = 0}^{N - 1} {\mathcal{B} \left( u_i \right)  + \mathcal{B} \left( \Delta_i  \right) },
\end{equation}
\begin{equation}
\mathcal{J}_f  = \left( {{\bf x}_N  - {\bf x}_r } \right)^{\rm T} {\bf Q}\left( {{\bf x}_N  - {\bf x}_r } \right) + \mathcal{B}\left( {\Delta _N } \right).
\end{equation}
\end{subequations}
Here,  the reference  state ${\bf x}_{r}$ = $\mathbf{0}$,  ${\bf Q}$/${\bf R}$ is the weighting matrix, and $\mathcal{B} \left( u_i \right)$ and $\mathcal{B} \left( \Delta_i  \right)$ are the corresponding barrier functions:
\begin{subequations}
\begin{equation}
\mathcal{B} \left( u_i \right) =  - \frac{1}{t}\left( {\log \left( {u_i  - \delta_{\min } } \right) + \log \left( {\delta_{\max }  - u_i } \right)} \right),
\end{equation}
\begin{equation}
\mathcal{B}\left( {\Delta _i } \right) = \left\{ \begin{array}{l}
\exp \left( {\Delta _i - \Delta _{i - 1} } \right)\quad\text{for}\quad \Delta _0 \ge 0, \\
\exp \left( {\Delta _{i - 1} - \Delta _i } \right)\quad\text{for}\quad \Delta _0 < 0, \\
\end{array} \right.
\end{equation}
\end{subequations}
where $\mathcal{B}$($u_i$) is used to limit control inputs and the high (low) steer bound is $\delta _{\max } \left( {\delta _{\min } } \right) = {\pi  \mathord{\left/
 {\vphantom {\pi  6}} \right.
 \kern-\nulldelimiterspace} 6}\left( { - {\pi  \mathord{\left/
 {\vphantom {\pi  6}} \right.
 \kern-\nulldelimiterspace} 6}} \right)$ rad. The objective of  $\mathcal{B} \left( \Delta _i \right)$ is to control the ego vehicle moving toward the lane center.

The first element of the optimal steering  sequence is then selected to define the normalized steering command at a given time  as
follows:\begin{equation}
{\rm SteerCmd} = \frac{{\delta _0^*  }}{{{\pi  \mathord{\left/
 {\vphantom {\pi  6}} \right.
 \kern-\nulldelimiterspace} 6}}}.
\end{equation}

\subsubsection{Longitudinal CILQR controller}
In the longitudinal direction, a proportional-integral (PI) controller \cite{Sam21} 
\begin{equation}
PI(v) = k_P e  + k_I \sum\limits_i {e_i } 
\end{equation}
is first applied to the ego car for tracking  reference speed $v_r$ under cruise conditions,  
where  $e=v-v_r$  and $k_P$/$k_I$ are  the  tracking error and the proportional/integral gain, respectively. The normalized acceleration command is then given as follows:  
\begin{equation}
{\rm AcclCmd} = \tanh (PI(v)).
\end{equation}
When a slower preceding vehicle is encountered, the AccelCmd must be updated to maintain a safe distance from that vehicle to avoid a collision; for this purpose, we use the following longitudinal CILQR algorithm.

The state variable and control  input for longitudinal inter-vehicle dynamics are defined as $
{\bf x'} = \left[ {\begin{array}{*{20}c}
   D & v & a  \\
\end{array}} \right]^{\rm T}
$ and $
{\bf u'} = \left[ j \right]$, respectively, where $a$, $j = \dot a$, and $D$ are the ego vehicle’s acceleration, jerk, and distance to the preceding car, respectively. The corresponding discrete-time system model is written as
\begin{equation}
{\bf x'}_{t + 1}  \equiv {\bf f'}\left( {{\bf x'}_t ,{\bf u'}_t } \right) = {\bf A'x'}_t  + {\bf B'u'}_t  + {\bf C'w'}, 
\end{equation}
where
\[
\begin{array}{l}
{\bf A'} = \left[ {\begin{array}{*{20}c}
1 & { - dt} & { - \frac{1}{2}dt^2 } \\
0 & 1 & {dt} \\
0 & 0 & 1 \\
\end{array}} \right],\quad{\bf B'} = \left[ {\begin{array}{*{20}c}
0 \\
0 \\
{dt} \\
\end{array}} \right], \\
{\bf C'} = \left[ {\begin{array}{*{20}c}
0 & {dt} & {\frac{1}{2}dt^2 } \\
0 & 0 & 0 \\
0 & 0 & 0 \\
\end{array}} \right],\quad{\bf w'} = \left[ {\begin{array}{*{20}c}
0 \\
{v_l } \\
{a_l } \\
\end{array}} \right]. \\
\end{array}
\]
Here, $v_l$/$a_l$ is the preceding car's speed/acceleration, and ${\bf w'}$ is the measurable disturbance input \cite{Qiu15}. The values of $D$ and $v_l$ are measured by the radar; $v$ is known; and $a = a_l  = 0$ is assumed. Here, MTUNets are used to recognize traffic objects,  and the radar is responsible for providing precise  distance measurements.

The objective function ($\mathcal{J}'$) for the longitudinal  CILQR controller can be written as,
\begin{subequations}
\begin{equation}
\mathcal{J}' = \mathcal{J}'_{ILQR}  + \mathcal{J}'_{b} + \mathcal{J}'_{f},
\end{equation}
\begin{equation}
\mathcal{J}'_{ILQR}  = \sum\limits_{i = 0}^{N - 1} {\left( {{\bf x'}_i  - {\bf x'}_r } \right)^{\rm T} {\bf Q'}\left( {{\bf x'}_i  - {\bf x'}_r } \right) + {\bf u'}_i^{\rm T} {\bf R'u'}_i }, 
\end{equation}
\begin{equation}
\mathcal{J}'_{b}  = \sum\limits_{i = 0}^{N - 1} {\mathcal{B}' \left( {u'_i} \right)  + \mathcal{B}' \left( D_i \right) + \mathcal{B}'\left( {a_i } \right)},
\end{equation}
\begin{equation}
\mathcal{J}_f'  = \left( {{\bf x'}_N  - {\bf x'}_r } \right)^{\rm T} {\bf Q'}\left( {{\bf x'}_N  - {\bf x'}_r } \right) + \mathcal{B'}\left( {D_N } \right)+ \mathcal{B'}\left( {a_N } \right).
\end{equation}
\end{subequations}
Here,  the reference state ${\bf x'}_r  = \left[ {\begin{array}{*{20}c}
   {D_r } & {v_l } & {a_l }  \\
\end{array}} \right]$, and $D_r$ is the reference distance for safety.  ${{\bf Q'}}$/${\bf R'}$ is the weighting matrix, and $\mathcal{B}' \left( {u'_i} \right)$, $\mathcal{B}' \left( D_i \right),$ and $\mathcal{B}'\left( {a_i } \right)$ are related  barrier functions:
\begin{subequations}
\begin{equation}
\mathcal{B}' \left( {u'_i} \right) =  - \frac{1}{t'}\left( {\log \left( {u'_i  - j_{\min } } \right) + \log \left( {j_{\max }  - u'_i } \right)} \right),
\end{equation}
\begin{equation}
\mathcal{B}' \left( D_i \right) = \exp \left( {D_r  - D_i } \right),
\end{equation}
\begin{equation}
\mathcal{B}'\left( {a_i } \right) =  \exp \left( {a_{\min }  - a_i } \right) + \exp \left( {a_i  - a_{\max } } \right),
\end{equation}
\end{subequations}
where $\mathcal{B}' \left( D_i  \right)$  is used for  maintaining a safe distance, and $\mathcal{B}'$($u'_i$) and $\mathcal{B}'$($a_i$) are used to limit the ego vehicle's jerk and acceleration to [$-$1, 1] m/s$^3$ and  [$-$5, 5] m/s$^2$, respectively.

The first element of the optimal jerk sequence is then chosen to update AccelCmd in the car-following scenario as 
\begin{equation}
{\rm AcclCmd} = \tanh \left( {PI\left( v \right)} \right) + j_0^*.
\end{equation}
The brake command (BrakeCmd) gradually increases in value from 0 to 1 when $D$ is smaller than a certain critical value during emergencies.

\subsection{The VPC algorithm}
The problematic scenario for the VPC algorithm is depicted in Fig. 3, which presents  a top-down view of fitted lane lines produced using our previous method \cite{Lee22}. First, the detected line segments [Fig. 3(a)] were clustered using the density-based spatial clustering of applications with noise (DBSCAN) algorithm. Second, the resulting semantic lanes were  transformed into BEV space by using a perspective transformation. Third, the least-squares quadratic polynomial fitting method was employed to produce  parallel ego-lane lines [Fig. 3(b)]; either of the two polynomials can be represented as $y=f(x)$. Fourth, the road curvature $\kappa$  was computed using the formula
\begin{equation}
\kappa  = \frac{{f''}}{{\left( {1 + f'^2 } \right)^{{3 \mathord{\left/
 {\vphantom {3 2}} \right.
 \kern-\nulldelimiterspace} 2}} }}.
\end{equation}
Because the curvature estimate from a single map is noisy, an average map obtained from eight consecutive frames was used for curve fitting. The resulting curvature estimates were then used to determine the correction value for the steering command in this study.

\begin{figure}[t]
\centerline{\includegraphics[scale=0.33]{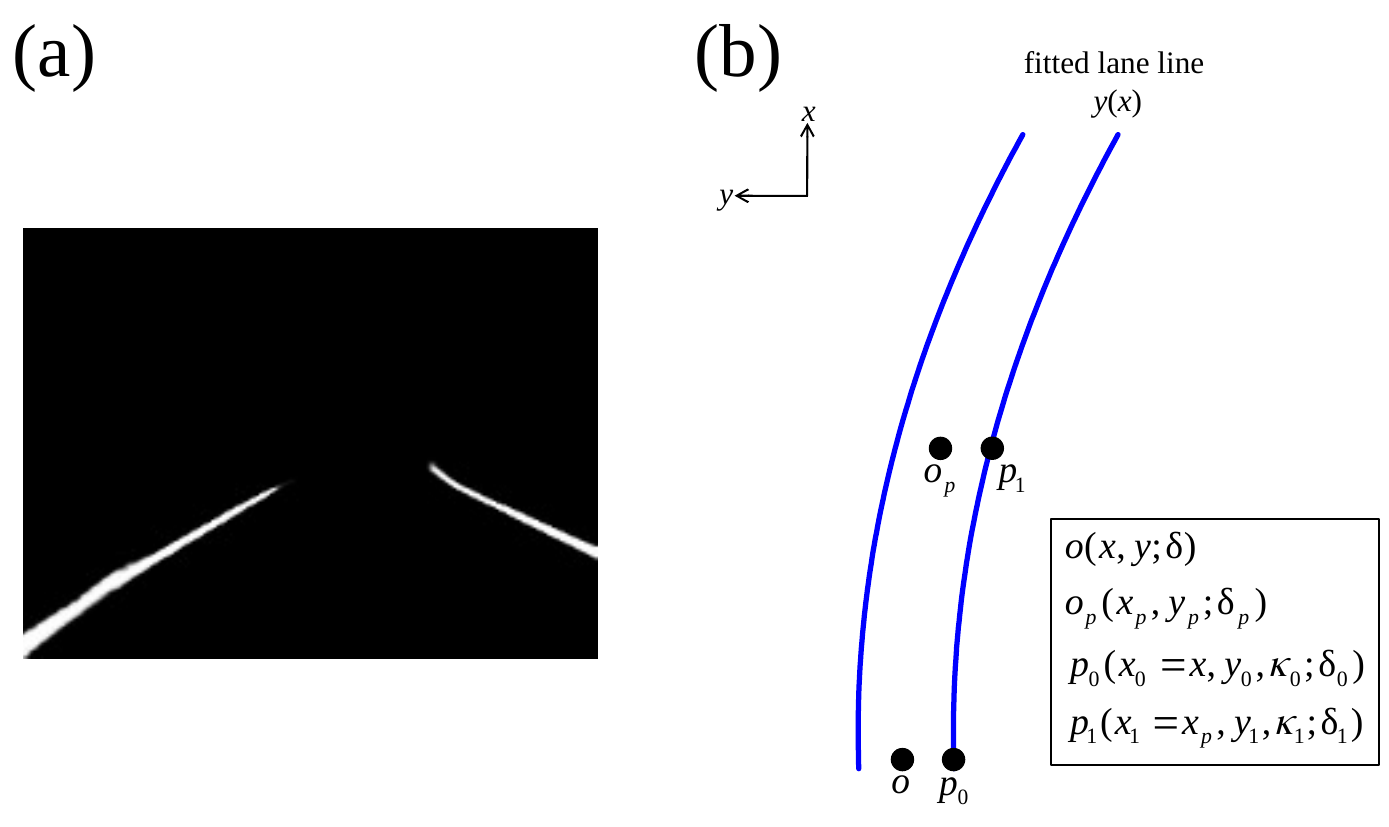}}
\caption{Problematic scenario for the VPC algorithm. (a) An example DNN-output lane-line binary map at a given time in the egocentric view. (b) Aerial view of the fitted lane lines. Here, $o$ is the current position of the ego vehicle and $o_p$ is the look-ahead point, and $p_0$  and $p_1$  represent the corresponding lane points at the same $x$ coordinates as $o$  and $o_p$, respectively. $\kappa$ and $\rm \delta$ are the road curvature and steering angle of the ego vehicle, respectively. In this paper, the  look-ahead distance $\overline {oo_p}$ = 10 m is used, which corresponds to a car speed of approximately 72 km/h \cite{Lee19}. }
\end{figure}

As shown in Fig. 3(b), $\delta$ at $o$ is the current steering angle. The desired steering angles at $p_0$  and $p_1$  can be computed using the local lane curvature \cite{Che19}:
\begin{subequations}
\begin{equation}
\delta _0  = \tan ^{ - 1}  \left( {c\kappa _0 } \right),
%\delta _0  = \arctan  \left( {c\kappa _0 } \right),
\end{equation}
\begin{equation}
\delta _1  = \tan ^{ - 1} \left( {c\kappa _1 } \right),
%\delta _1  = \arctan \left( {c\kappa _1 } \right),
\end{equation}
\end{subequations}
where $c$ is an adjustable parameter. Hence, the predicted steering angle at a look-ahead point $o_p$  can be represented as
\begin{equation}
{\delta _p} = \delta  + \left( {\delta _1  - \delta _0 } \right) \equiv \delta  + \Delta \delta.
\end{equation} 
Compared with those in existing LQR-based preview control methods \cite{ Zha19, Xu20}, fewer tuning parameters are required when the VPC algorithm is included in the steering geometry model; moreover, the algorithm can be combined with other path-tracking models. For example, a VPC-CILQR controller can update the CILQR steering command [Eq. (22)] as follows:
\begin{equation}
\begin{array}{l}
 {\rm VPC\_SteerCmd} \\ 
  = \left\{ \begin{array}{l}
 {\rm SteerCmd} + \left| {\Delta \delta } \right|{\rm \quad if \quad SteerCmd} \ge {\rm 0,} \\ 
 {\rm SteerCmd} - \left| {\Delta \delta } \right|{\rm \quad if \quad SteerCmd} < {\rm 0}{\rm .} \\ 
 \end{array} \right. \\ 
 \end{array}
\end{equation}

\begin{figure}[!t]
\centerline{\includegraphics[scale=0.5]{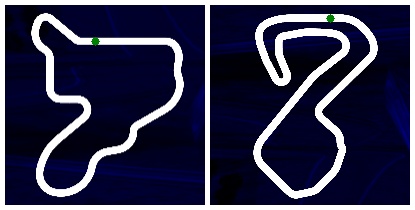}}
\caption{Tracks A (left) and B (right) for dynamically evaluating proposed MTUNet and control models. The  total length of Track A/B (Track 7/8 in \cite{Lee21a}) was 2843/3919 m with lane width 4 m, and the maximum curvature was approximately 0.03/0.05 1/m, which was curvier than a typical road \cite{Fit94}. The self-driving car drove in a counterclockwise direction, and the starting locations are marked by green filled circle symbols. A self-driving vehicle \cite{Li19} could not finish a lap on Track A using the direct perception approach  \cite{Che15}.}
\end{figure}

In summary, the proposed VPC algorithm uses the future road curvature at a look-ahead point 10 m in front of the ego car (Fig. 3) as input to generate the updated steering inputs. This algorithm is applied before the ego car enters a curvy road to improve tracking performance. Accurate and complete future road shape prediction is crucial for developing preview path-tracking control algorithms \cite{Xu20}. However, whether the necessary information can be obtained is greatly dependent on the maximum perception range of lane detection modules. As demonstrated in Fig. 5, LLAMAS \cite{Beh19} data are more useful than TORCS \cite{Lee21a} or CULane \cite{Pan18} datasets for developing algorithms with such path-tracking functionality. A nonlinear MPC approach using high-quality predicted lane curvature data can achieve better control performance over the proposed method; however, if computational cost is a concern, such a nonlinear approach may not necessarily be preferred. The following sections describe validation experiments where the proposed algorithm was compared against other control algorithms.

\begin{table*}[!t]
\caption{Datasets used in the experiments}
\begin{center}
\begin{tabular}{l|c|c|c|c|c}
\hline
Dataset &  Scenarios &  No. of images&  Labels  & No. of traffic objects & Sources   \\ \hline
\multirow{2}{*}{CULane} & urban, & \multirow{2}{*}{28368} &  & \multirow{2}{*}{80437} & \cite{Pan18}, \\
                  &        highway           &                   &         ego-lane lines,          &                   &       this work     \\ \cline{1-3} \cline{5-6} 
\multirow{2}{*}{LLAMAS} & \multirow{6}{*}{highway} & \multirow{2}{*}{22714} &       bounding boxes            & \multirow{2}{*}{29442} & \cite{Beh19}, \\
                  &                   &                   &                   &                   &        this work           \\ \cline{1-1} \cline{3-6} 
\multirow{4}{*}{TORCS} &                   & \multirow{4}{*}{42747} & ego-lane lines, & \multirow{4}{*}{30189} & \multirow{4}{*}{\cite{Lee21a}} \\
                  &                   &                   &         bounding boxes,           &                   &                   \\
                  &                   &                   &          ego's heading,         &                   &                   \\
                  &                   &                   &         road type          &                   &                   \\ \hline
\end{tabular}
\end{center}
\end{table*}

\begin{figure*}[t]
\centerline{\includegraphics[scale=0.2]{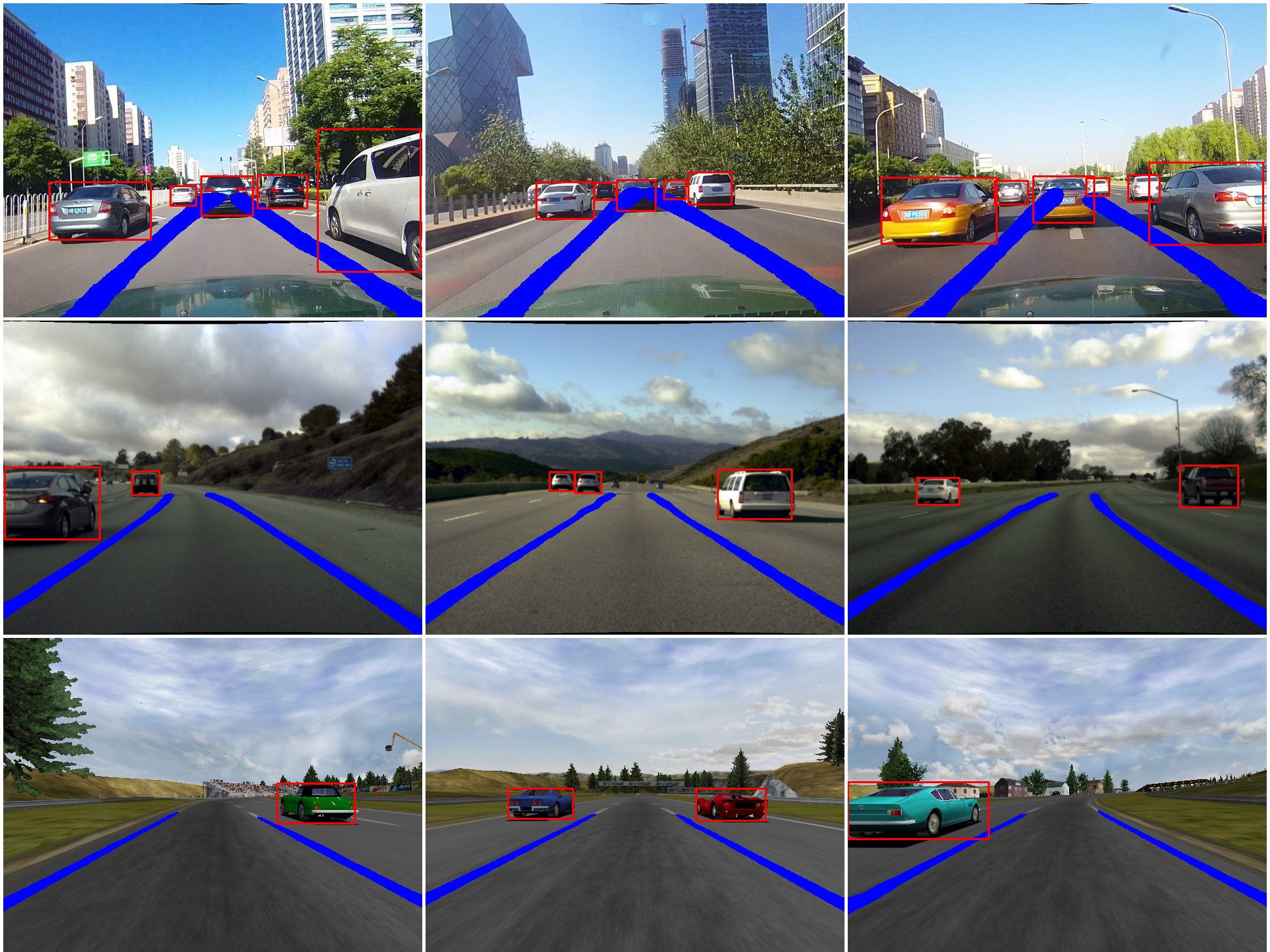}}
\caption{Example traffic object and lane-line detection results for the MTUNet$\_$1$\times$ network on  CULane (first row), LLAMAS (second row), and TORCS (third row) images.}
\end{figure*}

\section{Experimental setup}
The proposed MTUNets extract local and global contexts from input images to simultaneously perform segmentation, detection, and pose tasks. Because the these tasks have different learning rates \cite{Bru18,Liu16,Red17}, the proposed MTUNets were trained in a stepwise instead of end-to-end manner to help the backbone network learn common features. The training strategy, image data, and validation are described as follows.

\subsection{Network training strategy}
The MTUNets were trained in three stages. The pose subnet was first trained through stochastic gradient descent (SGD) with a batch size (bs) of 20, momentum (mo) of 0.9, and learning rate (lr) starting from $10^{ - 2}$ and decreasing by a factor of 0.9 every 5 epochs for a total of 100 epochs. The detection and pose subnets were then trained jointly with the parameters obtained in the first training stage and using the SGD optimizer with bs = 4, mo = 0.9, and lr = $10^{ - 3}$, $10^{ - 4}$, and $10^{ - 5}$ for the first 60 epochs, the 61st to 80th epochs, and the last 20 epochs, respectively. All subnets (detection, pose, and segmentation) were trained together in the last stage with the  pretrained  model obtained in the previous stage and using the Adam optimizer. Bs and mo were set to 1 and 0.9, respectively, and lr was set to $10^{ - 4}$ for the first 75 epochs and $10^{ - 5}$ for the last 25 epochs. The total loss in each stage was a weighted sum of the corresponding losses \cite{Lee17}.

\subsection{Image datasets}
We conducted experiments on the artificial TORCS \cite{Lee21a} and real-world CULane \cite{Pan18} and LLAMAS \cite{Beh19} datasets. The summary statistics of the datasets are presented in Table  IV. The customized TORCS dataset has joint labels for all tasks, whereas the original CULane/LLAMAS dataset only contained lane line labels. Thus, we annotated each CULane and LLAMAS image with traffic object bounding boxes to mimic the TORCS dataset. Correspondingly, the TORCS, CULane, and LLAMAS datasets had approximately 30 K, 80 K, and 29 K labeled traffic objects, respectively. To determine anchor boxes for the detection task, the $k$-means algorithm \cite{Mac67} was applied to partition the ground truth boxes. The CULane and LLAMAS datasets lack ego vehicle angle labels; therefore, these datasets could only be used for evaluations in segmentation and detection tasks. The ratio of the number of images used in the training phase to that used in the test phase was approximately 10 for all datasets, as in our previous works \cite{Lee21a, Lee22}. Recall/average precision (AP; IoU was set to 0.5), recall/F1 score, and accuracy/mean absolute error (MAE) were used to evaluate model performance in detection, segmentation, and pose tasks, respectively.

\begin{table*}[!t]
\caption{Performance of trained MTUNets on the test data}
\begin{center}
\begin{tabular}{l|c|c|c|c|c|c|c|c}
%\resizebox{\textwidth}{!}{\begin{tabular}{l|c|c|c|c|c|c|c|c}
\hline
\multirow{3}{*}{Network} & \multirow{3}{*}{Dataset} & \multirow{3}{*}{Task config.}  &\multicolumn{2}{c|}{Det}  &\multicolumn{2}{c|}{Seg} & \multicolumn{2}{c}{Pose} \\ \cline{4-9}
& &  & \multirow{2}{*}{Recall} & \multirow{2}{*}{AP (\%)} & \multirow{2}{*}{Recall} & \multirow{2}{*}{F1 score} & Heading & Road type \\ 
& &  &         &  &                 &                           & MAE (rad)     & accuracy (\%)  \\ \hline
MTUNet$\_$2$\times$ & \multirow{3}{*}{CULane}  & \multirow{3}{*}{Det + Seg} & 0.858& 65.32  & 0.694 & 0.688  & - & - \\ 
MTUNet$\_$1$\times$ &   & &  0.831  & 58.67  & 0.658 & 0.595  & - & - \\ 
MTMResUNet  &   & &  0.852  & 54.28  &  0.559 & 0.568  & - & - \\  \hline

MTUNet$\_$2$\times$ & \multirow{3}{*}{LLAMAS}  & \multirow{3}{*}{Det + Seg} &0.942& 64.42 &0.935& 0.827 & - & - \\ 
MTUNet$\_$1$\times$ &   & & 0.946  &  59.96& 0.936& 0.831 & - & - \\ 
MTMResUNet  &   & & 0.950  & 57.40 & 0.736 & 0.748 & - & - \\  \hline

MTUNet$\_$2$\times$ & \multirow{9}{*}{TORCS}  & \multirow{3}{*}{Pose} &-& - &-& - & 0.004 & 90.42 \\ 
MTUNet$\_$1$\times$ &   &  &-& - &-& - & 0.005 & 90.48 \\ 
MTMResUNet &   &   &-& - &-& -& 0.006 & 83.77 \\ \cline{3-9}\cline{1-1}

MTUNet$\_$2$\times$ &   & \multirow{3}{*}{Det + Seg} &0.976& 71.51 &0.905& 0.889 & - & - \\ 
MTUNet$\_$1$\times$ &   &  &0.974& 66.14 &0.904& 0.894 & - & - \\ 
MTMResUNet &   &   &0.968& 66.12 &0.833& 0.869 & - & - \\ \cline{3-9}\cline{1-1}
MTUNet$\_$2$\times$ &   & \multirow{3}{*}{Det + Seg + Pose} &0.952& 65.83 &0.922& 0.883 & 0.005 & 87.08 \\ 
MTUNet$\_$1$\times$ &   &  &0.956& 59.25 &0.901& 0.882 & 0.004 & 94.30 \\ 
MTMResUNet &   &   &0.959& 51.88 &0.830& 0.855 & 0.007 & 80.46 \\ 
\hline
\end{tabular}
%\end{tabular}}
\end{center}
\end{table*}

\subsection{Autonomous driving simulation}
The open-source driving environment TORCS provides sophisticated physics and graphics engines; it is therefore ideal for not only visual processing but also vehicle dynamics research \cite{Wym00}. The ego vehicle controlled by our self-driving framework was driven autonomously on unseen TORCS roads [e.g., Tracks A and B in Fig. 4] to validate the effectiveness of our approach. All experiments, including both MTUNet training and testing and driving simulations, were conducted  on a PC equipped with an INTEL i9-9900K CPU, 64 GB of RAM, and an NVIDIA RTX 2080 Ti GPU with 4352 CUDA cores and 11 GB of GDDR memory. The control frequency for the ego vehicle in TORCS was approximately 150 Hz on this computer.

\begin{table}[!t]
\caption{Results for MTUNets in terms of parameters (Params), MACs, and FPS}
\begin{center}
\begin{tabular}{l|c|c|c|c}
\hline
Network  & Task config. & Params & MACs & FPS \\ \hline
MTUNet$\_$$2\times$ & \multirow{3}{*}{Pose} &  19.37 M & 13.93 G &  74.02  \\ 
MTUNet$\_$$1\times$  & & 4.98 M & 3.51 G & 122.72   \\
MTMResUNet & & 5.37 M  & 2.70 G & 99.08  \\ \hline
MTUNet$\_$$2\times$ & \multirow{3}{*}{Det + Seg} & 68.62  M & 47.36 G &  24.44 \\ 
MTUNet$\_$$1\times$  & & 21.70 M & 12.89 G & 43.58  \\
MTMResUNet & & 21.97 M  & 15.98 G & 27.79  \\ \hline
MTUNet$\_$$2\times$ & \multirow{3}{*}{Det + Seg + Pose} &  83.31 M & 50.55 G &  23.28  \\ 
MTUNet$\_$$1\times$  & &  25.50 M & 13.69 G & 40.77  \\
MTMResUNet & & 26.56 M  & 16.95 G & 27.30  \\ \hline
\end{tabular}
\end{center}
\end{table}

\begin{table}[!t]
\caption{Dynamic system models and parameters for implementing the CILQR and SQP controllers}
\begin{center}
\begin{tabular}{l|c|c}
\hline
  & Lateral & Longitudinal \\ 
  & CILQR-/SQP-based & CILQR/SQP \\ \hline
Dynamic model    & Eq. (19) & Eq. (25) \\ 
Sampling time ($dt$)  & 0.05 s  & 0.1 s\\ 
Pred. horizon ($N$) & 30  & 30 \\ 
Ref. dist. ($D_r$) & -  &  11 m  \\ 
Weighting matrixes  & ${\bf Q}$ = ${\rm diag}\left( {20 ,1 ,20 ,1 } \right)$  & ${{\bf Q'}}$ = ${\rm diag}\left( {20 ,20 ,1 } \right)$ \\ 
  & ${\bf R} = \left[ 1 \right]$ &${\bf R'} = \left[ 1 \right]$\\ \hline
\end{tabular}
\end{center}
\end{table}

\begin{table*}[!h]
\caption{Performance of the VPC-CILQR, CILQR, VPC-SQP, and SQP algorithms with MTUNet$\_$1$\times $ in terms of the MAE for the tests in Figs. 6--14}
\begin{center}
\begin{tabular}{c|c|c|cc|cc|cc|cc}
\hline
Maneuver  &Track &Speed (km/h)& \multicolumn{2}{c|}{VPC-CILQR} & \multicolumn{2}{c|}{CILQR}& \multicolumn{2}{c|}{VPC-SQP} & \multicolumn{2}{c}{SQP} \\\hline
 \multirow{5}{*}{Lane-keeping} &  &  & $\theta$ (rad) & $\Delta $ (m) & $\theta$ (rad) & $\Delta $ (m)& $\theta$ (rad) & $\Delta $ (m) & $\theta$ (rad) & $\Delta $ (m) \\ 
  &A &76  &  0.0086  &  0.0980  &  0.0083 &  0.1058 &  0.0122 &  0.1071   &  0.0118 & 0.1187  \\ 
 &A & 80   &  0.0099    &   0.1091   &  0.0098   &  0.1097   &  -  &   -   &   -  &   -  \\
  &B &50  & 0.0079  &  0.0748   & 0.0074  & 0.0775  &  0.0099 &  0.1286   & 0.0121  &  0.1470 \\ 
 &B & 60  &  0.0078   &   0.0779  &  0.0079  &  0.0783   &  0.0099  &   0.1083   &  0.0112   & 0.1147   \\ 
\hline
\multirow{2}{*}{Car-following} & & & $v$ (m/s) & $D$ (m) & $v$ (m/s) & $D$ (m)& $v$ (m/s) & $D$ (m) & $v$ (m/s) & $D$ (m) \\ 
 &B$^{a}$ & -     &  - &   -  & 0.1971  &  0.4201 &  - &   -  & 0.2629  &  0.4930 \\ \hline
\multicolumn{3}{c|}{Related trajectories}   & \multicolumn{2}{c|}{Figs. 6, 10} & \multicolumn{2}{c|}{Figs. 7, 11, 14}& \multicolumn{2}{c|}{Figs. 8, 12} & \multicolumn{2}{c}{Figs. 9, 13, 14} \\\hline
 
\multicolumn{4}{l}{$^{a}$\scriptsize{Computation  from 1150 to 1550 m}} 
\end{tabular}
\end{center}
\end{table*}

\begin{table*}[!t]
\caption{Average computation time of VPC, CILQR, and SQP algorithms}
\begin{center}
\begin{tabular}{c|c|c|c}
\hline
Task &VPC & CILQR & SQP \\ \hline
Lane-keeping& 15.56 ms  & 0.58 ms & \multicolumn{1}{r}{9.70 ms}\\ 
Car-following & - & 0.65 ms & 14.01 ms \\  \hline
\end{tabular}
\end{center}
\end{table*}

\section{Results and discussions}
Table  V presents the performance results of the MTUNet models on the testing data for various tasks. Table VI lists the number of parameters, computational complexity, and inference speed of each scheme as a comparison of computational efficiency. As described in Section II, although the input size of the MTUNet models was reduced by the use of padded 3 $\times$ 3 Conv layers, model performance was not affected; MTUNet$\_$2$\times$/MTUNet$\_$1$\times$ achieved similar results to our previous model in the segmentation and pose tasks on the TORCS and LLAMAS datasets \cite{Lee21a}. For complex CULane data, the MTUNet model performance performed worse than the SCNN \cite{Pan18}, the original state-of-the-art method for this dataset; however, the SCNN had lower inference speed because of its higher computational complexity \cite{Don21}. The MTUNet models are designed  for real-time control of self-driving vehicles; the SCNN model is not. Of the three considered MTUNet variants, MTUNet$\_$2$\times$ and MTUNet$\_$1$\times$ outperformed  MTMResUNet on all datasets if each model jointly performed the detection and segmentation tasks (first, second, and fourth row of Table V). This result differs from that of a previous study on a single-segmentation task for biomedical images \cite{Nab20}. Task gradient interference can reduce the performance of an MTDNN  \cite{Sta20, Kok17}; in this case, the MTUNet$\_$2$\times$ and MTUNet$\_$1$\times$ models outperformed the complex  MTMResUNet network because of their elegant architecture. When the pose task was included (last row of Table V), MTUNet$\_$2$\times$ and MTUNet$\_$1$\times$ also outperformed MTMResUNet on all evaluation metrics; the decreasing AP scores for the detection task are attributable to an increase in false positive (FP) detections. However, for all models, the inclusion of the pose task only decreased the recall scores for the detection task by approximately 0.02 (last two rows of Table V); nearly 95$\%$ of the ground truth boxes were still detected during when the models simultaneously performed all tasks. Following the method for efficiency analysis used in \cite{Bia18} (Sec. V. B. in \cite{Bia18}), this study computed the densities of the detection AP and road type accuracy scores using the data in the last row of Table V/VI. MTUNet$\_$1$\times$ had higher efficiency in terms of parameter utilization than did MTUNet$\_$2$\times$. MTUNet$\_$1$\times$ was 3.26 times smaller than MTUNet$\_$2$\times$ and achieved a 1.75 times faster inference speed (40.77 FPS); this speed is comparable to that of the YOLOP model \cite{Don21}. These results indicate that MTUNet$\_$1$\times$ is the most efficient model for collaborating with controllers to achieve automated driving. The MTUNet$\_$1$\times$ model can also be run on a low-performance computer with only a few gigabytes of GDDR memory. For a computer with a GTX 1050 Max-Q GPU  with  640 CUDA cores and 4 GB of GDDR memory, the MTUNet$\_$1$\times$ model achieved an inference speed of 14.69 FPS for multi-task prediction. Fig. 5 presents example MTUNet$\_$1$\times$ network outputs for both traffic objects and lane detection on all datasets.

\begin{figure}[!t]
\centerline{\includegraphics[scale=0.3]{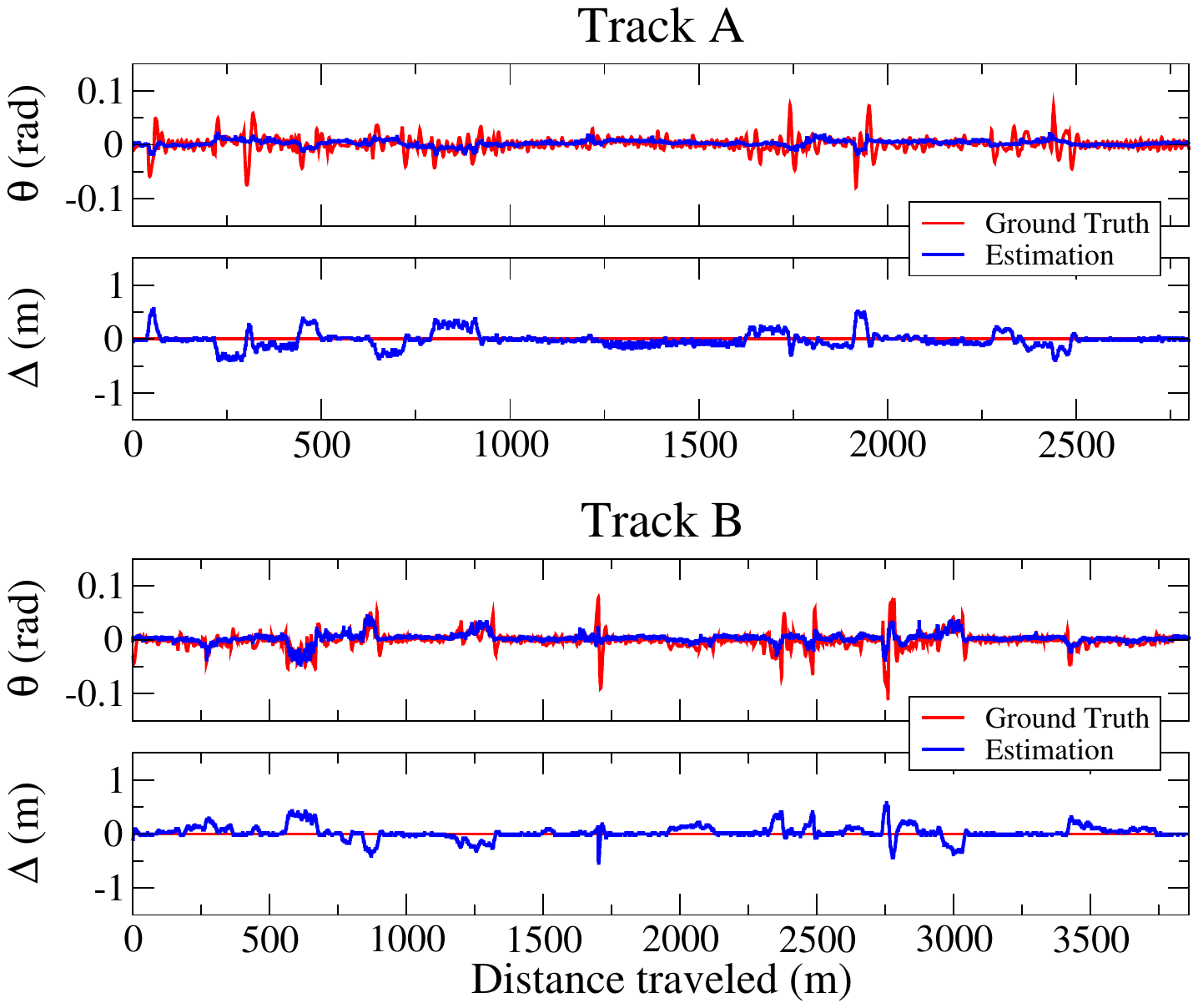}}
\caption{Dynamic performance of lateral VPC-CILQR algorithm and MTUNet$\_$1$\times $ model for an ego vehicle with heading  $\theta$ and lateral offset  $\Delta$ for lane-keeping maneuvers in the central lanes of Tracks A and B at 76 and 50 km/h, respectively. At the curviest section of Track A (near 1900 m), the maximal $\Delta$ value was 0.52 m; the ego car controlled by this model outperformed the ego car controlled by the Stanley controller (Fig. 3 in \cite{Lee21a}), MTL-RL [Fig. 11(a) in \cite{Li19}], or CILQR (Fig. 7) algorithms.}
\end{figure}

\begin{figure}[!h]
\centerline{\includegraphics[scale=0.3]{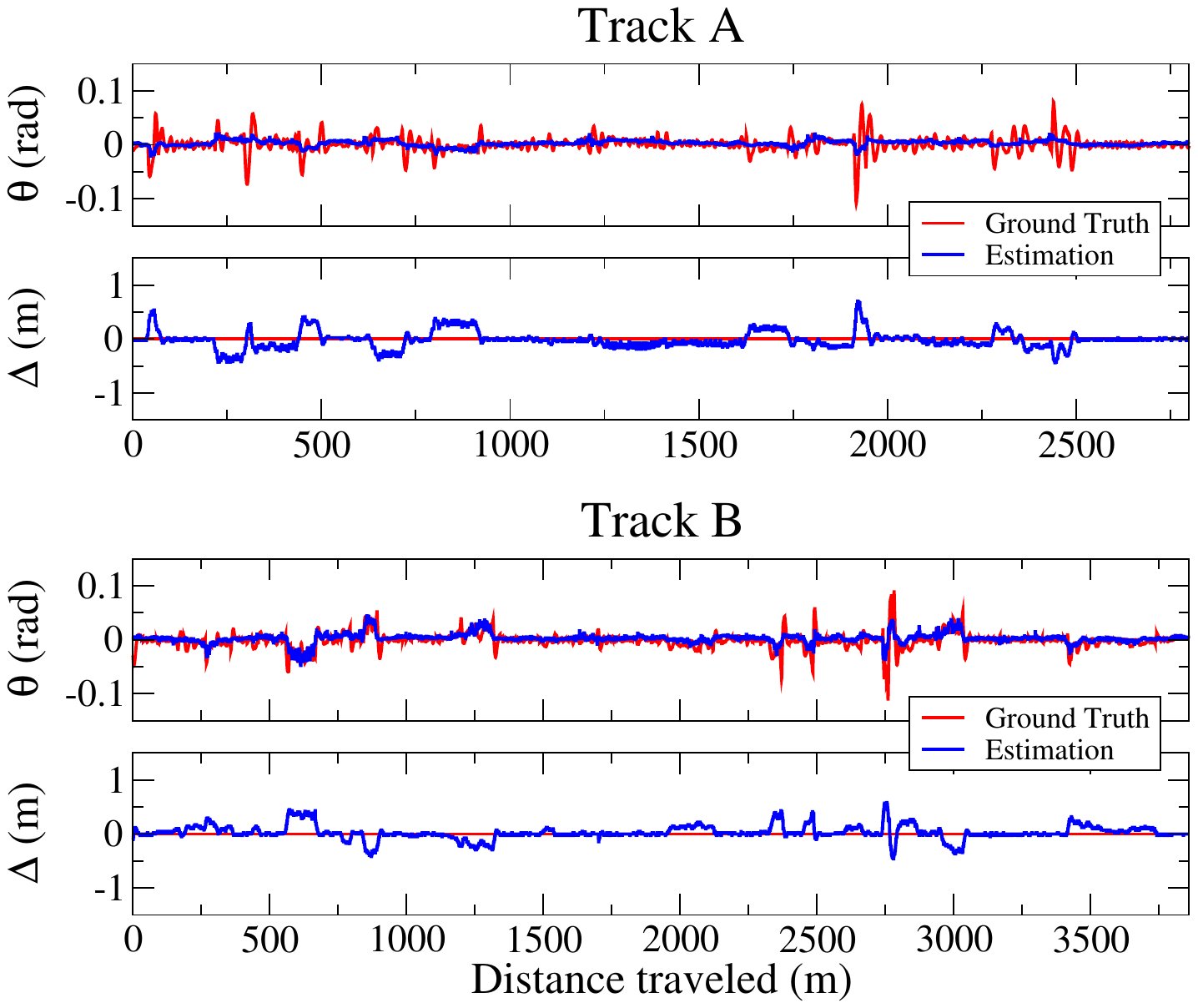}}
\caption{Dynamic performance of the lateral CILQR algorithm and MTUNet$\_$1$\times $ model for lane-keeping maneuvers in the central lanes of Tracks A and B at the same speeds as those in Fig. 6. At the curviest section of Track A (near 1900 m), the maximal $\Delta$ value was 0.71 m, which is 1.36 times larger than that of the ego car controlled by the VPC-CILQR algorithm.}
\end{figure}

\begin{figure}[!t]
\centerline{\includegraphics[scale=0.3]{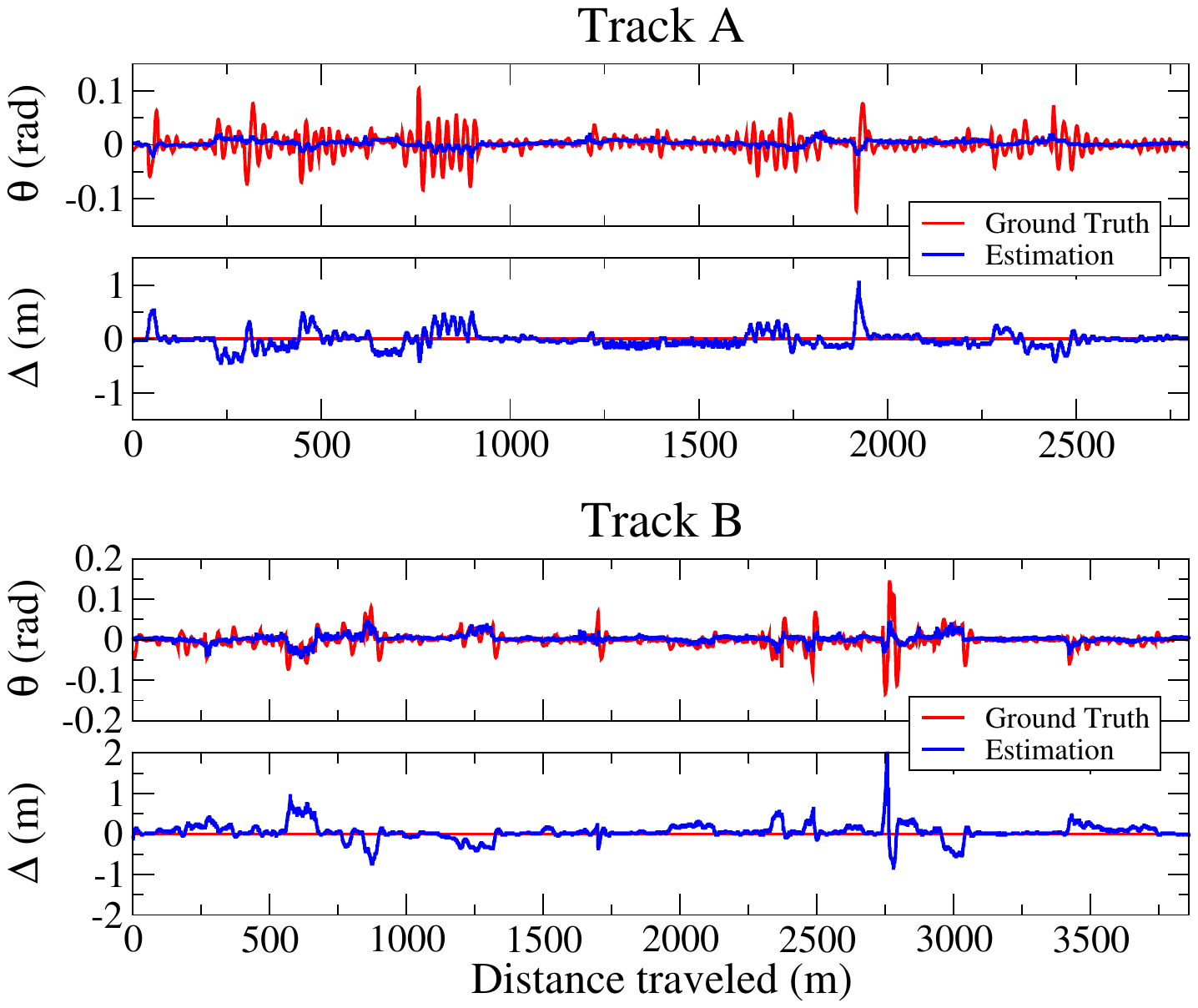}}
\caption{Dynamic performance of the lateral VPC-SQP algorithm and MTUNet$\_$1$\times $ model for lane-keeping maneuvers in the central lanes of Tracks A and B at the same speeds as those in Fig. 6. For the curviest sections of Tracks A and B  (near  1900 and 2750 m, respectively),  the performance of the VPC-SQP algorithm was inferior to those of the VPC-CILQR and CILQR algorithms (Figs. 6 and 7). These algorithms also outperformed the SQP algorithm (Fig. 9), indicating the effectiveness of the VPC algorithm. }
\end{figure}

\begin{figure}[!t]
\centerline{\includegraphics[scale=0.3]{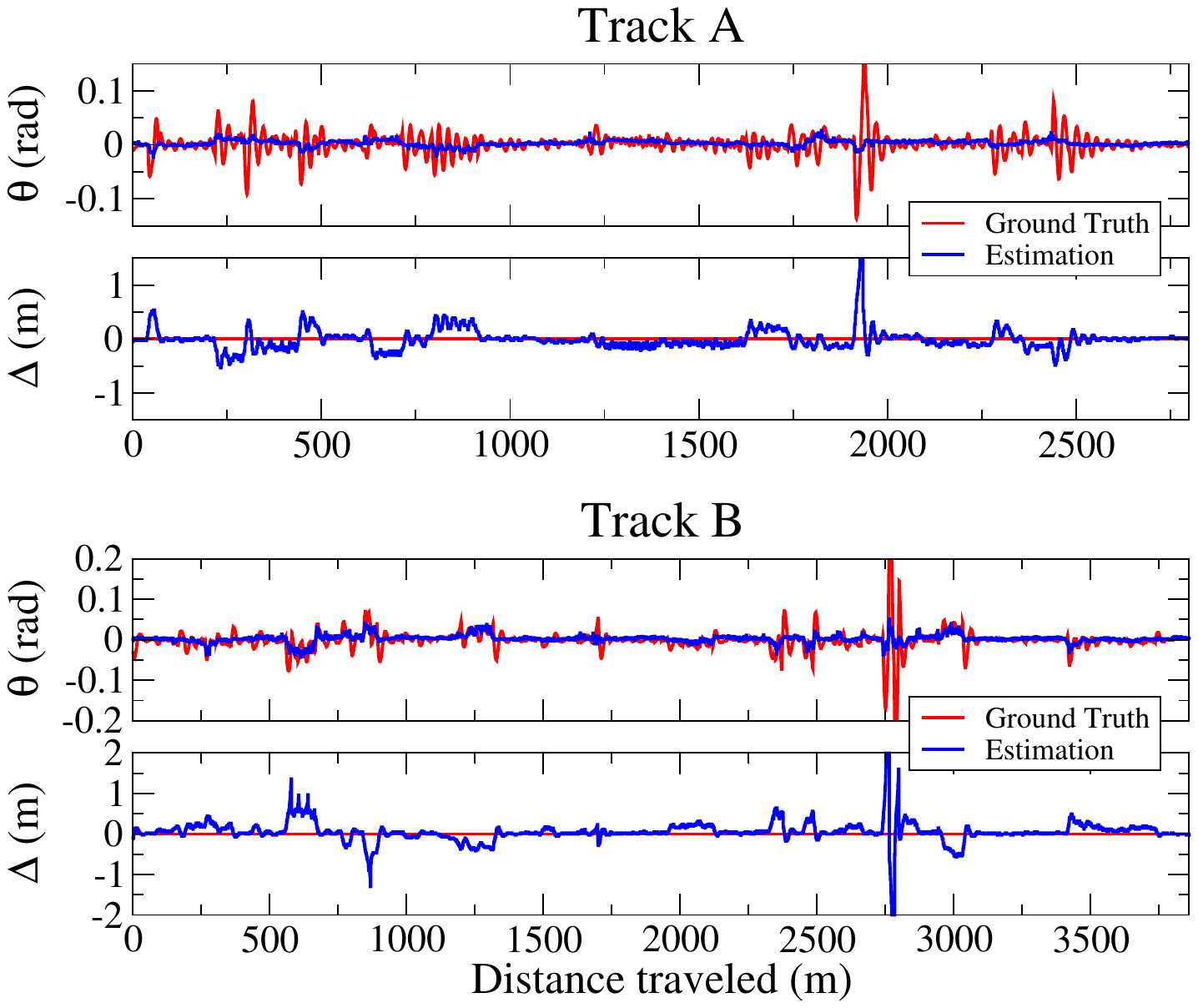}}
\caption{Dynamic performance of the lateral SQP algorithm and MTUNet$\_$1$\times $ model for lane-keeping maneuvers in the central lanes of  Tracks A and B at the same speeds as those in Fig. 6. The model performance at the curviest section of Tracks A and B  (near  1900 and 2750 m, respectively) was inferior to those of all other tested methods.}
\end{figure}

\begin{figure}[!t]
\centerline{\includegraphics[scale=0.3]{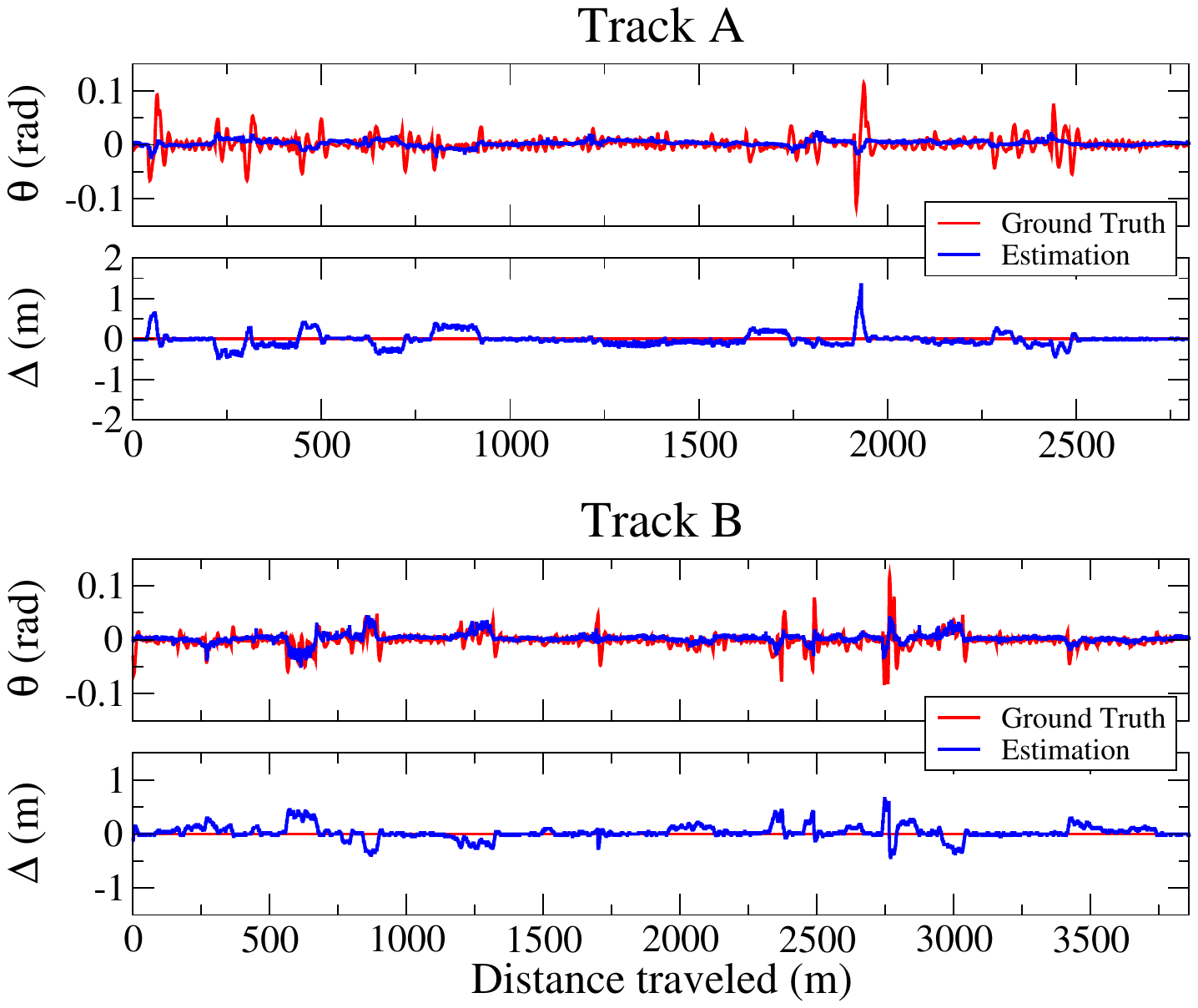}}
\caption{Dynamic performance of the lateral VPC-CILQR algorithm and MTUNet$\_$1$\times $ model for an ego vehicle with heading  $\theta$ and lateral offset  $\Delta$ for lane-keeping maneuvers in the central lanes of Tracks A and B at 80 and 60 km/h, respectively. At the curviest section of Track A, the maximal $\Delta$ value was 1.34 m.}
\end{figure}

\begin{figure}[!t]
\centerline{\includegraphics[scale=0.3]{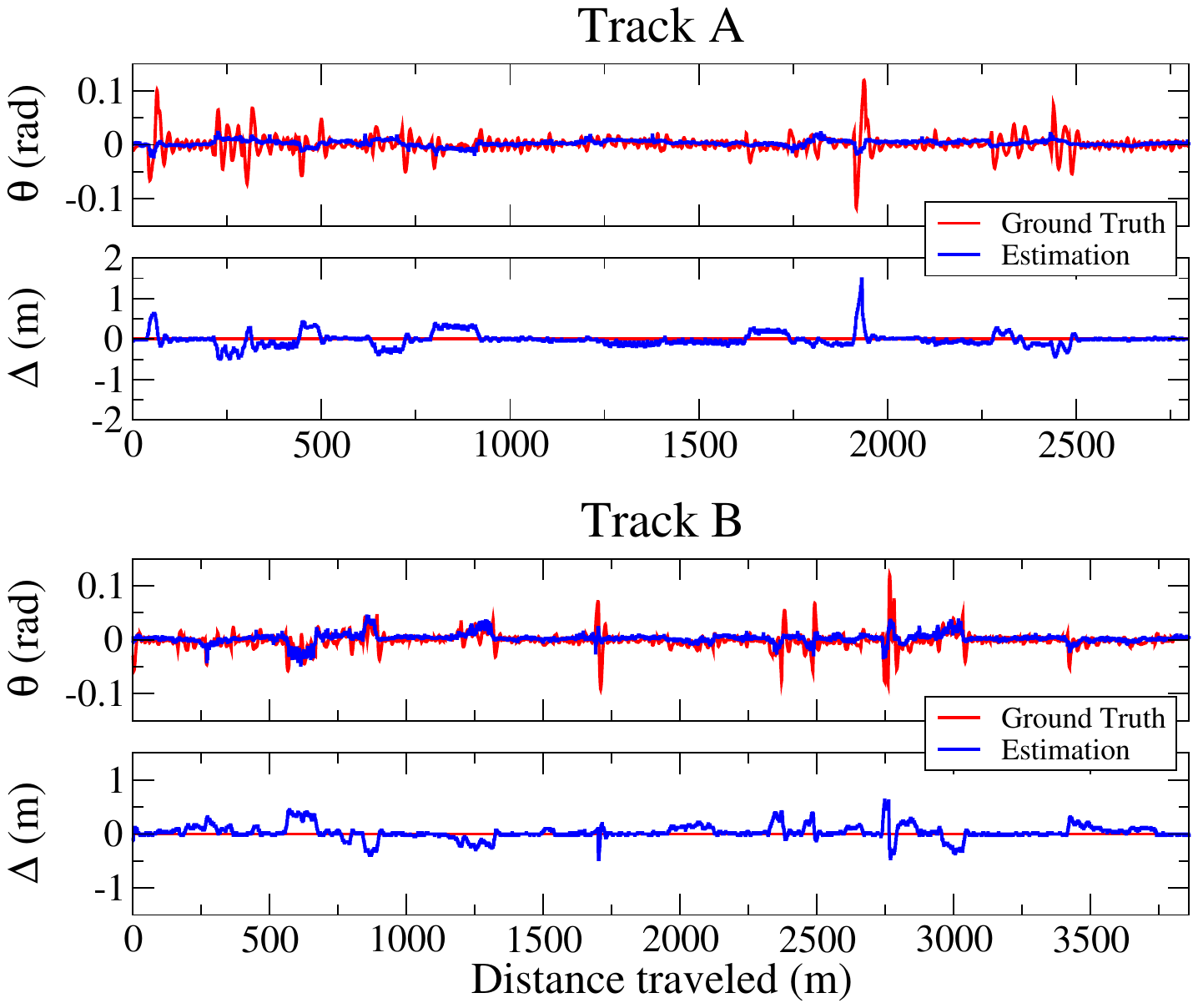}}
\caption{Dynamic performance of the lateral CILQR algorithm and MTUNet$\_$1$\times $ model for lane-keeping maneuvers in the  central lanes of Tracks A and B at the same speeds as those in Fig. 10. At the curviest section of Track A, the maximal $\Delta$ value was 1.48 m.}
\end{figure}

\begin{figure}[!t]
\centerline{\includegraphics[scale=0.3]{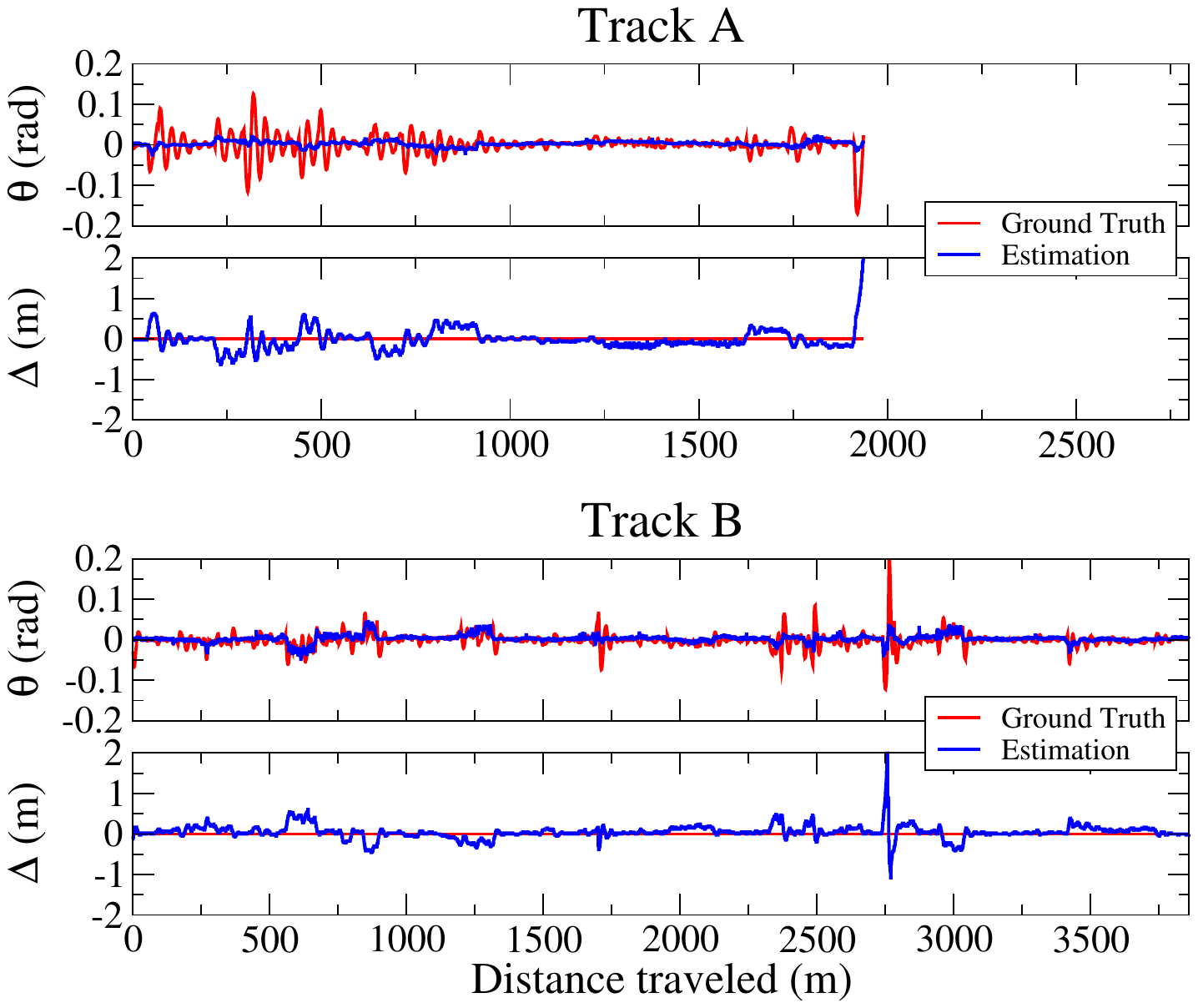}}
\caption{Dynamic performance of the lateral VPC-SQP algorithm and MTUNet$\_$1$\times $ model for lane-keeping maneuvers in the central lanes of Tracks A and B at the same speeds as those in Fig. 10.}
\end{figure}

\begin{figure}[!t]
\centerline{\includegraphics[scale=0.3]{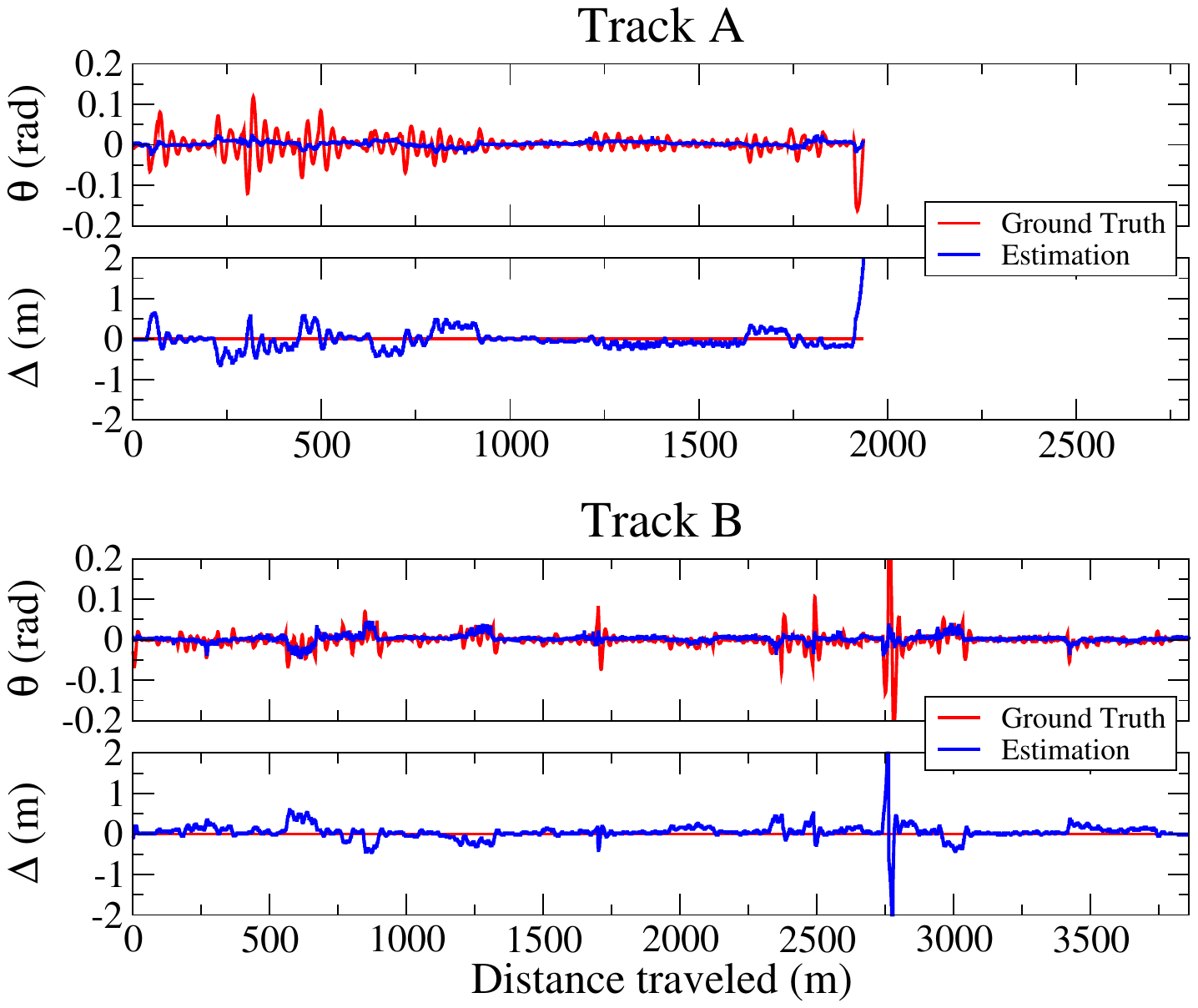}}
\caption{Dynamic performance of the lateral SQP algorithm and MTUNet$\_$1$\times $ model for lane-keeping maneuvers in the  central lanes of Tracks A and B at the same speeds as those in Fig. 10.}
\end{figure}

To objectively evaluate the dynamic performance of the autonomous driving algorithms, lane-keeping and car-following maneuvers were performed on the challenging tracks, Tracks A and B, as shown in Fig. 4. The SQP-based controllers were implemented using the ACADO toolkit \cite{Hou11} for comparison with the CILQR-based controllers. All settings for these algorithms were the same as summarized in Table  VII. For the lateral control experiments, autonomous vehicles were designed to drive at various cruise speeds on Tracks A and B. The $\theta$ and $\Delta$ results for the VPC-CILQR, CILQR, VPC-SQP, and SQP algorithms are presented in Figs. 6--13. The results of the CILQR and SQP controllers for the longitudinal control experiments are presented in Fig. 14. The MAEs for $\theta$, $\Delta$, $v$, and $D$ in Figs. 6--14 are listed in Table  VIII. Table  IX presents the average time to arrive at a solution for the VPC, CILQR, and SQP algorithms. The inference time was shorter for VPC than MTUNet$\_$1$\times$ (24.52 ms). Moreover, the CILQR had a computation speed that was faster than the ego vehicle control period (6.66 ms); the SQP solvers were slower. Specifically, the computation time per cycle for the lane-keeping and car-following tasks, respectively, for the SQP solvers were 16.7 and 21.5 times longer than those of the CILQR solvers. A discussion of the results for all the tested controllers are presented as follows.

\begin{figure}[!t]
\centerline{\includegraphics[scale=0.3]{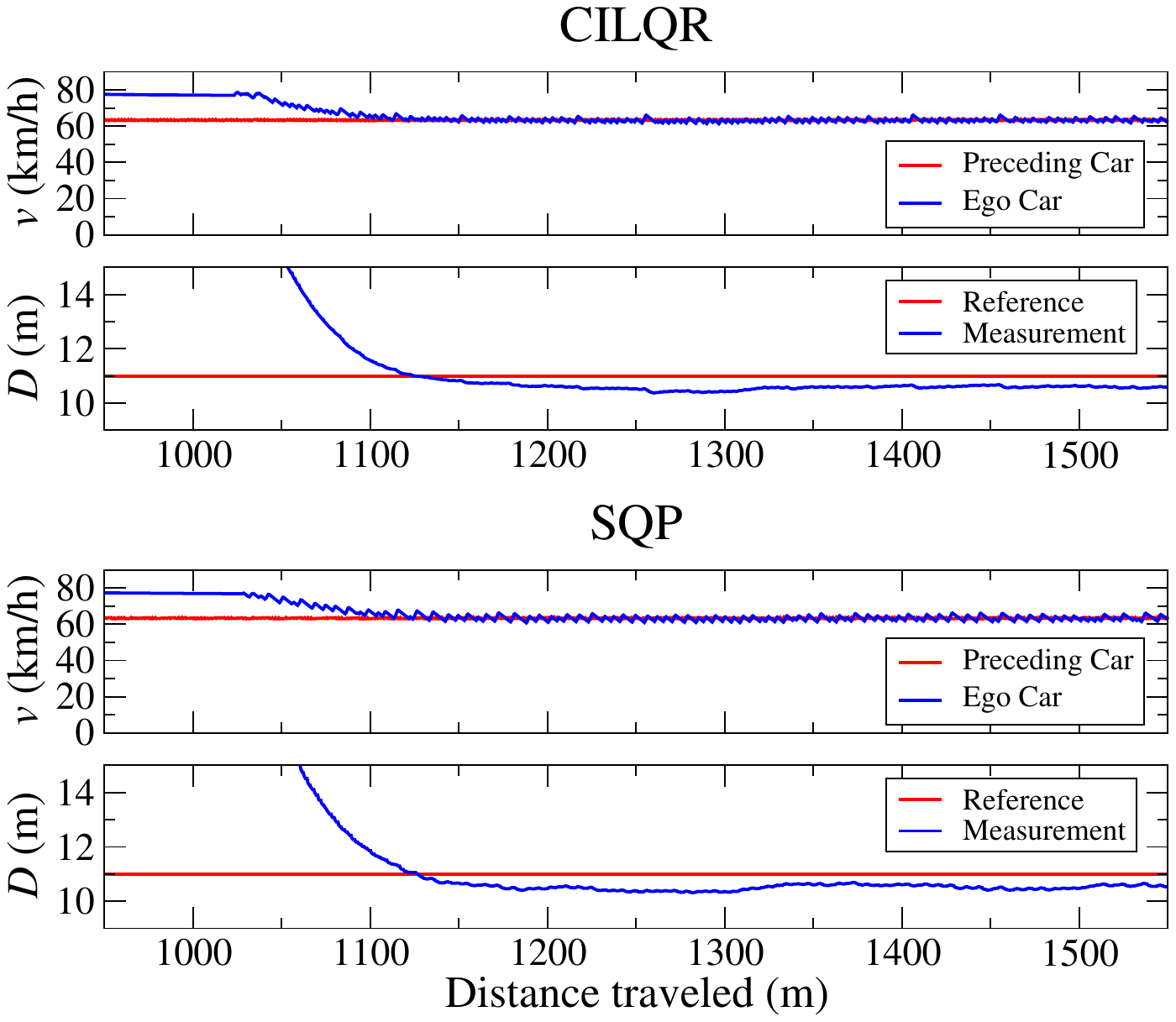}}
\caption{Results for the longitudinal CILQR and SQP algorithms in car-following scenario after ego car travels 1075 m on Track B; $v$ and  $D$ are speed and intervehicle distance, respectively.}
\end{figure}

In Figs. 6--9, all methods, including the MTUNet$\_$1$\times$ model, could effectively guide the ego car to drive along the lane center to complete one lap at cruise speeds of 76 and 50 km/h on Tracks A and B, respectively. The discrepancy in the  $\theta$  between the MTUNet$\_$1$\times$ estimation and the ground truth trajectory was attributable to curvy or shadowy road segments, which may induce vehicle jittering \cite{Li19}. Nevertheless, the $\Delta$ values estimated from lane line segmentation were more robust in difficult scenarios than those obtained with the end-to-end method \cite{Che15}. Therefore, these $\Delta$ values can be used by controllers to effectively correct $\theta$ errors and return the ego car to the road’s center. The maximum  $\Delta$ deviations from the ideal zero values on Track A (g-track-3 in \cite{Li19}) were smaller when the ego car was controlled by the VPC-CILQR controller (shown in Fig. 6) than when it was controlled by the CILQR \cite{Che19} (Fig. 7) or MTL-RL \cite{Li19} algorithms. Note that the vehicle speed in the MTL-RL control framework on Track A  for that study was 75 km/h, which is slower than that in this study. This finding indicates that for curvy roads, the VPC-CILQR algorithm better minimized $\Delta$ than did the other investigated algorithms. Due to the lower computation efficiency of the standard SQP solver \cite{Che19}, the SQP-based controllers were less effective for maintaining the ego vehicle's stability than the CILQR-based controllers on the curviest sections of Tracks A and B (Figs. 8 and 9). Moreover, the VPC-SQP algorithm outperformed the SQP algorithm alone, further demonstrating the effectiveness of the VPC algorithm. In terms of MAE, the VPC-CILQR controller outperformed the other methods in terms of $\Delta$-MAE on both tracks (data for 76 and 50 km/h in Table VIII). However, $\theta$-MAE was 0.0003 and 0.0005 rad  higher on Tracks A and B, respectively, for the VPC-CILQR controller than the CILQR controller. This may have been because the optimality of CILQR solution is losing if the external VPC algorithm is applied to it. This problem could be solved by applying standard MPC methods with more general lane-keeping dynamics, such as the lateral control model presented in \cite{Xu20}, which  uses road curvature to describe vehicle states. This nonlinear MPC design is computationally expensive and may not meet the requirements for real-time autonomous driving.

In Figs. 10--13, the ego car was guided to drive along the central lane by the MTUNet$\_$1$\times$ model at higher cruise speeds (80 and 60 km/h on Tracks A and B, respectively) than those in Figs. 6--9. For ego vehicles with the VPC-CILQR and CILQR controllers (Figs. 10 and 11), the maximum $\Delta$ deviations were approximately half of the lane width (2 m). In particular, the ego cars controlled by the SQP-based algorithms unintentionally left the ego-lane at the curviest section of Track A (Figs. 12 and 13). This was attributed to the slower reaction times of SQP-based algorithms (9.70 ms) than of CILQR-based algorithms (0.58 ms). Therefore, higher controller latency may not only result in ego car instability but also unsafe driving, particularly when the vehicle enters a curvy road at high speed.

The car-following maneuver in Fig. 14 was performed on a section of Track B. The ego vehicle was initially cruising  at 76 km/h  and approached a slower preceding car with speed in the range of 63 to 64 km/h. For all ego vehicles with the CILQR or SQP controllers, the vehicle speed was regulated, the preceding vehicle was tracked, and the controller maintained a safe distance between the vehicles. However, the uncertainty in the optimal solution led to differences between the reference and response trajectories \cite{Lim22}. For the longitudinal CILQR and SQP controllers, respectively, $v$-MAE was 0.1971 and 0.2629 m/s, and  $D$-MAE was 0.4201 and 0.4930 m (second row of Table VIII). Hence, CILQR again outperformed SQP in this experiment. A supplementary video featuring the lane-keeping and car-following simulations can be found at \url{https://youtu.be/Un-IJtCw83Q}.

\section{Conclusion}
In this study, a vision-based self-driving system that uses a monocular camera and radars to collect sensing data is proposed; the system comprises an MTUNet network for environment perception and VPC and CILQR modules for motion planning. The proposed MTUNet model is an improvement on our previous model \cite{Lee21a}; we have added a YOLOv4 detector and increased the network’s efficiency by reducing the network input size for use with TORCS \cite{Lee21a}, CULane \cite{Pan18}, and LLAMAS \cite{Beh19} data. The most efficient MTUNet model, namely  MTUNet$\_$1$\times $, achieved an inference speed of 40.77 FPS for simultaneous lane line segmentation, ego vehicle pose estimation, and traffic object detection tasks. For vehicular automation, a lateral VPC-CILQR controller was designed that can plan vehicle motion based on the ego vehicle’s heading, lateral offset, and road curvature as determined by MTUNet$\_$1$\times $ and postprocessing methods. The longitudinal CILQR controller is activated when a slower preceding car is detected. The optimal jerk is then applied to regulate the ego vehicle’s speed to prevent a collision. The  MTUNet$\_$1$\times$ and VPC-CILQR controller can collaborate for ego vehicle operation on challenging tracks in TORCS; this algorithm outperforms methods based on the CILQR \cite{Che19} or MTL-RL \cite{Li19} algorithms for the same path-tracking task on the same large-curvature roads. Moreover, the self-driving vehicle with long-latency SQP-based controllers tended to leave the lane on some curvy routes. By contrast, the short-latency CILQR-based controllers could drive stably and safely in the same scenarios. In conclusion, the experiments demonstrated the applicability and feasibility of the proposed system, which comprises perception, planning, and control algorithms. It is suitable for real-time autonomous vehicle control and does not require HD maps. A future study can apply the proposed  autonomous driving system to a real vehicle operating on actual roads.

%\begin{IEEEbiography}[{\includegraphics[width=1in,height=1.25in,clip,keepaspectratio]{DerHauLee.jpg}}]{Der-Hau Lee}
%\end{IEEEbiography}

%\begin{IEEEbiographynophoto}{Der-Hau Lee}
%received his Ph.D. degree in Physics from the Department of Electrophysics, National Yang Ming
%Chiao Tung University, Taiwan, in 2018. His research interests include machine learning and autonomous driving.
%\end{IEEEbiographynophoto}

\end{document}